\documentclass[10pt,final]{IEEEtran}
\usepackage{nopageno}
\usepackage{graphicx}
\usepackage{epsfig}
\usepackage{amsmath}
\usepackage{bbm}
\usepackage{amsthm} 
\usepackage{amssymb}
\usepackage{setspace}
\usepackage{wrapfig}
\usepackage{times,color}
\usepackage{caption}
\usepackage{subcaption}
\usepackage{algorithm}
\usepackage{algpseudocode}
\usepackage{pifont}
\usepackage{epstopdf}
\usepackage{cite,footnote,xspace,syntonly,bm,balance}
\usepackage{amsfonts}

\theoremstyle{definition}
\input{mysymbol.sty}

\allowdisplaybreaks[2]
\algnewcommand{\Inputs}[1]{%
  \State \textbf{Inputs:}
  \Statex \hspace*{\algorithmicindent}\parbox[t]{.8\linewidth}{\raggedright #1}
}
\algnewcommand{\Initialize}[1]{%
  \State \textbf{Initialize:}
  \Statex \hspace*{\algorithmicindent}\parbox[t]{.8\linewidth}{\raggedright #1}
}

\def \bb {{\bf b}}

\def \bbeta {{\boldsymbol \beta}}

\def\baselinestretch{1.03}
 \long\def\symbolfootnote[#1]#2{\begingroup
 	\def\thefootnote{\fnsymbol{footnote}}
 	\footnote[#1]{#2}\endgroup} \psfull
 \makeatother

\begin{document}

\title{\huge Graph-adaptive Nonlinear Dimensionality Reduction}
\author{\IEEEauthorblockN{Yanning Shen,~\IEEEmembership{Student
			Member,~IEEE,} Panagiotis A. Traganitis,~\IEEEmembership{Student
			Member,~IEEE,}\\
		and  Georgios B. Giannakis,~\IEEEmembership{Fellow,~IEEE}}\\
	\thanks{Yanning Shen, Panagiotis A. Traganitis and Georgios B. Giannakis are
		with the Dept.\ of Electrical and Computer Engineering and the
		Digital Technology Center, University of Minnesota,
		Minneapolis, MN 55455, USA.\protect\\   
		This work was supported by the NSF grants 171141, 1500713, 1442686 and NIH 1R01GM104975-01.\protect\\
		E-mails: shenx513@umn.edu, traga003@umn.edu, georgios@umn.edu}
}
\maketitle  




\begin{abstract}
	In this era of data deluge, many signal processing and machine learning tasks are faced with high-dimensional datasets, including images, videos, as well as time series generated from social, commercial and brain network interactions. Their efficient processing calls for dimensionality reduction techniques capable of properly compressing the data while preserving task-related characteristics, going beyond pairwise data correlations. The present paper puts forth a nonlinear dimensionality reduction framework that accounts for data lying on known graphs. The novel framework encompasses most of the existing dimensionality reduction methods, but it is also capable of capturing and preserving possibly nonlinear correlations that are ignored by linear methods. Furthermore, it can take into account information from multiple graphs. The proposed algorithms were tested on synthetic as well as real datasets to corroborate their effectiveness.
\end{abstract}

\begin{keywords}
	Dimensionality reduction, nonlinear modeling, signal processing over graphs
\end{keywords}


\vspace{-1.5mm}

\section{Introduction}
\label{sec:intro}
The massive development of connected devices and highly precise instruments has introduced the world to vast volumes of high-dimensional data. Traditional data analytics cannot cope with these massive amounts, which motivates well investigating dimensionality reduction schemes capable of gleaning out efficiently low-dimensional information from large-scale datasets. Dimensionality reduction is a vital first step to render tractable critical learning tasks, such as large-scale regression, classification, and clustering of high-dimensional datasets. In addition, dimensionality reduction can allow for accurate visualization of high-dimensional datasets.

Dimensionality reduction methods have been extensively studied by the signal processing and machine learning communities~\cite{jolliffe2002principal,roweis2000nonlinear,belkin2003laplacian,scholkopf1997kernel}. Principal component analysis (PCA)~\cite{jolliffe2002principal} is the `workhorse' method yielding low-dimensional representations that preserve most of the high-dimensional data variance. Multi-dimensional scaling (MDS)~\cite{kruskal1978multidimensional} on the other hand, maintains pairwise distances between data when going from high- to low-dimensional spaces, while local linear embedding (LLE)~\cite{roweis2000nonlinear} only preserves relationships between neighboring data. Information from non-neighboring data is lost in LLE's low-dimensional representation, which may in turn influence the performance of ensuing tasks such as classification or clustering~\cite{suykens1999least,hartigan1979algorithm}. It is also worth stressing that all aforementioned approaches capture and preserve linear dependencies among data. However, for data residing on highly nonlinear manifolds using only linear relations might produce low-dimensional representations that are not accurate. Generalizing PCA, kernel PCA (KPCA) can capture nonlinear relationships between data, for a preselected kernel function. In addition, Laplacian eigenmaps~\cite{belkin2003laplacian} preserve nonlinear similarities between neighboring data.

While all the aforementioned approaches have been successful in reducing the dimensionality of various types of data, they do not consider additional information during the dimensionality reduction process. This prior information may be task specific, e.g. provided by some ``expert'' or by the physics of the problem, or it could be inferred from alternative views of the data, and can provide additional insights for the desired properties of the low-dimensional representations. In fMRI signals for instance, in addition to time series collected at different brain regions, one may also have access to the structural connectivity patterns among these regions. 

At the same time, data may arrive from multiple heterogeneous sources, e.g. in addition to fMRI time courses, electroencephalography time series might be available. While it is desirable to draw inferences from all these multimodal data, their heterogeneous nature inhibits the use of traditional statistical learning tools. Thus, schemes that can generate useful data representations by fusing judiciously the information contained in different data modes are required.

 As shown in \cite{shahid2016fast,jin2015low,shang2012graph,jiang2013graph} for PCA, useful additional information can be encoded in a graph, and incorporated into the dimensionality reduction process through \emph{graph-adaptive} regularization.
	PCA accounting for the graph Laplacian has been advocated in \cite{jiang2013graph}, to improve performance by exploiting the underlying graph structure. A low rank matrix factorization method incorporating multiple graph regularizers for linear PCA can be found in \cite{jin2015low}. However, a quadratic program must be solved per iteration to optimally combine the adopted graph regularizers. Robust versions of linear graph PCA have also been reported\cite{shahid2016fast}. Multiple graph regularizers were also studied in \cite{wang2013multiple} relying on low-rank matrix matrix factorization. 
\\
\noindent\textbf{Our contributions.}
The present manuscript presents a novel \emph{graph-adaptive} (GRAD)  \emph{nonlinear} dimensionality reduction approach, to account for prior information  on one or multiple graphs. By extending the concept of kernel PCA to graphs, our approach encompasses all aforementioned approaches, while markedly broadening their scope.
Compared to our conference precursor in \cite{shen2017camsap}, the present manuscript includes GRAD nonlinear dimensionality reduction when domain knowledge is unknown. To this end, a multi-kernel based approach is developed that uses the data to select the appropriate kernel for the dimensionality reduction task.
In addition, we show how our approach can reduce dimensionality of multi-modal datasets, by considering separate graphs induced by different modes. Further, we generalize our approach to semi-supervised scenarios, where labels for a few data are available.

The rest of the paper is organized as
follows. Section~\ref{sec:pre} provides preliminaries along
with notation and background works.
Section~\ref{sec:graph} introduces the proposed 
GRAD nonlinear dimensionality reduction scheme, while Section~\ref{sec:general}
provides pertinent generalizations and applications. Section~\ref{sec:numerical_tests} presents numerical tests
conducted to evaluate the performance of the novel dimensionality reduction scheme. Finally, concluding remarks and future
research directions are given in Section~\ref{sec:conclusion}. 

\noindent\textbf{Notation:} Unless otherwise noted, lowercase bold letters $\bm{x}$ denote
vectors, uppercase bold letters $\mathbf{X}$ represent matrices, and
calligraphic uppercase letters $\mathcal{X}$ stand for sets. The
$(i,j)$th entry of  $\mathbf{X}$ is denoted by
$[\mathbf{X}]_{ij}$; $\mathbf{X}^{\top}$ denotes the transpose of  $\bbX$, while $\bbX^{\dagger}$ denotes the Moore-Penrose pseudo-inverse of matrix $\bbX$. The
$D$-dimensional real Euclidean space is denoted by $\mathbb{R}^{D}$,  the set of positive real numbers by $\mathbb{R}_{+}$, the positive integers by $\mathbb{Z}_{+}$, and the $\ell_2$-norm by $\|\cdot\|$.

\section{Preliminaries and Problem Statement}\label{sec:pre}


 Consider a dataset with $N$ vectors of dimension $D$ collected as columns of the matrix $\bbY:=[\bby_1, \dots, \bby_N]$. Without loss of generality, it will be assumed that the data are centered, that is the sample mean $N^{-1}\sum_{i=1}^N \bby_i$ has been removed from each $\bby_i$. For future use,  the singular value decomposition (SVD) of the data matrix $\bbY$ is $\bbY = \bbU\bbSigma\bbV^{\top}$. Dimensionality reduction seeks a set of $d<D$-dimensional vectors $\{\bbpsi_i\}_{i=1}^{N}$, that preserve certain properties of the original data $\{\bby_i\}_{i=1}^N$. 

The following subsections review popular dimensionality reduction schemes, that can be viewed as special cases of kernel PCA. 
 
 
 \subsection{Principal component analysis }
 Given data $\bbY$, PCA finds a linear subspace of dimension $d$ such that all the data lie on or close to it, in the Euclidean distance sense. Specifically, PCA solves
 \begin{align}
 \label{eq:pca}
\min_{\bbU_d,\{\bbpsi_i\}}\sum_{i=1}^N\|\bby_i-\bbU_d\bbpsi_i\|_2^2~~~~ \text{s. to   }~~ \bbU_d^\top\bbU_d=\bbI
 \end{align}
where $\bbU_d\in \mathbb{R}^{D\times d}$ is an orthonormal matrix whose columns span the sought subspace. 
The optimal solution of \eqref{eq:pca} is ${\bbpsi}_i={\bbU}_d^\top\bby_i$, where ${\bbU}_d$ is formed by the eigenvectors of $\bbY\bbY^\top = \mathbf{U}\mathbf{\Sigma}^2\mathbf{U}^{\top}$ corresponding to the $d$  eigenvalues with the largest magnitude, or equivalently to the $d$ leading left singular vectors of $\bbY$~\cite{trevor2011elements}. Given $\{\bbpsi_i\}$, the original vectors can be recovered as $\hat{\bby}_i = {\bbU}_d\bbpsi_i$. PCA has well-documented merits when data lie close to a $d$-dimensional hyperplane. Its complexity is that of eigendecomposing $\bbY\bbY^\top$, i.e., $\mathcal{O}(ND^2)$, which means PCA is more affordable when $D\ll N$. In contrast, dimensionality reduction of small sets of high-dimensional vectors $(D\gg N)$ becomes more tractable with the dual PCA that we outline next.

\subsection{Dual PCA and Kernel PCA}
Collect all the lower dimensional data representations as columns of the $d\times N$ matrix $\bbPsi:=[\bbpsi_1,\dots, \bbpsi_N]$. Then using the SVD of $\bbY$, we find 
\begin{align}
\label{eq:dPCA2}
\bbPsi=\bbU_d^\top \bbY=\bbSigma_d \bbV_d^\top
\end{align}
where $\bbSigma_d\in\mathbb{R}^{d\times d}$ is a diagonal matrix containing the $d$ leading singular values of $\bbY$, and $\bbV_d\in \mathbb{R}^{N\times d}$ is the submatrix of $\bbV$ collecting the corresponding right singular vectors of $\bbY$. Since $\bbY^{\top}\bbY = \mathbf{V}\mathbf{\Sigma}^2\mathbf{V}^{\top}$, the low-dimensional representations of data can be obtained through the eigendecomposition of $\bbY^{\top}\bbY$. Using this method to find the low-dimensional representations of the data is known as \emph{Dual PCA}.
  As only eigendecomposition of $\bbY^{\top} \bbY$ is required, the complexity of dual PCA is $\mathcal{O}(DN^2)$; therefore, it is preferable when $D\gg N$. Moreover, it can be readily verified that besides \eqref{eq:pca}, $\bbPsi$ is the optimal solution to the following optimization problem (see Appendix~\ref{app:kernel})
\begin{align}
\label{eq:dPCA3}
	&\min_{\bbPsi}\|\bbK_y -\bbPsi^\top\bbPsi\|_F^2
	~~~~\text{s. to }~\bbPsi\bbPsi^\top=\bbLambda_d
\end{align}
%
 where $\bbK_y:=\bbY^\top\bbY$ is known as the Gram or kernel matrix, and $\bbLambda_d$ denotes a $d\times d$ diagonal matrix  containing the $d$ largest eigenvalues of $\bbK_y$. Compared to PCA, dual PCA requires only the inner products $\{\bby_i^\top\bby_j\}$ in order to obtain the low-dimensional representations. Hence, dual PCA can yield low-dimensional vectors $\{\bbpsi_i\}$ of general (non-metric) objects that are not necessarily expressed using vectors $\{\bby_i\}$, so long as inner products (meaning correlations) of the latter are known. On the other hand, the original data $\{\bm{y}_i\}$ cannot be recovered from $\{\bm{\psi}_i\}$ found by the solution of \eqref{eq:dPCA3}. 
 
Consider now expanding the cost in \eqref{eq:dPCA3}, to equivalently express it as
 \begin{align}
 \label{eq:dPCA4}
 	&\min_{\bbPsi:\bbPsi\bbPsi^\top={\small\bbLambda_d}} {\rm tr}(\bbPsi\bbK_y^{-1}\bbPsi^\top)
 	\end{align}
 
 	Recalling that ${\bbK}_y^{-1}$ is symmetric and nonnegative definite with eigenvalues equal to the inverses of the eigenvalues of ${\bbK}_y$, we can re-write \eqref{eq:dPCA4} as
 	\begin{align}
 	\label{eq:dPCA5}
 	 \min_{\bbPsi:\bbPsi\bbPsi^\top=\bbLambda_d} -{\rm tr}(\bbPsi\bbK_y\bbPsi^\top) 
 \end{align}
 where now $\bbLambda_d$ contains the $d$ largest eigenvalues of $\bbK_y$.

\begin{table}[htb]
		\begin{center}
			\begin{tabular}{|p{2.6cm} | p{3.2cm} | p{1.6cm}|}
				\hline
				\textbf{Kernel type}  & $\kappa(\bby_i,\bby_j)$  & \textbf{Parameters} \\
				\hline
				\hline
				Linear    & $\bby_i^{\top}\bby_j$ & -  \\
				\hline
				Gaussian     &
				$\exp\{\frac{-\|\bby_i - \bby_j\|_2^2}{2\sigma^2}\}$  & 
				$\sigma> 0$
				\\
				\hline
				Polynomial kernel    & $(\bby_i^{\top}\bby_j + c)^{p}$ & $p>0$, $c$  \\
				\hline
			\end{tabular}
		\end{center}
		\caption{Examples of kernels.}
		\label{tab:kernelexamples}
	\end{table} 
  
 While PCA performs well for data that lie close to a hyperplane, this property might not hold for the available data $\bbY$~\cite{jin2015low}. In such cases one may resort to kernel PCA. Kernel PCA ``lifts'' $\{\bby_i\}$ using a nonlinear function $\bbphi$, onto a higher (possibly infinite) dimensional space, where the data may lie on or near a linear hyperplane, and then finds low-dimensional representations $\{\bbpsi_i \}$. Kernel PCA is obtained by solving \eqref{eq:dPCA3} or \eqref{eq:dPCA4} with
$[\bbK_y]_{i,j}=\kappa(\bby_i,\bby_j) = \bbphi^{\top}(\bby_i)\bbphi(\bby_j)$, where $\kappa(\bby_i,\bby_j)$ denotes a prescribed kernel function \cite{ham2004kernel}. Table~\ref{tab:kernelexamples} lists a few popular kernels used in the literature, including the linear kernel which links linear dual PCA with kernel PCA. 

	\subsection{Local linear embedding}
Another popular method that deals with data that cannot be presumed  close to a hyperplane is local linear embedding (LLE)~\cite{roweis2000nonlinear}. LLE postulates that $\{\bby_i\}$ lie on a smooth manifold, which can be locally approximated by tangential hyperplanes. Specifically, LLE assumes that each datum can be expressed as a linear combination of its neighbors; that is, 
$
\bby_i=\sum_{j\in\mathcal{N}_i} w_{ij}\bby_j+\bbe_i \label{eq:lle:vec}
$,
where $\mathcal{N}_i$ is a set containing the indices of the nearest neighbors of $\bby_i$, in the Euclidean distance sense, and $\bbe_i$ captures unmodeled dynamics.

In order to solve for $\{w_{ij}\}$, the following optimization problem is considered
\begin{align}
\label{eq:lle_kx}
&{\bbW} =\arg\min_{\check{\bbW}}\|\bbY-\bbY\check{\bbW}\|_F^2\nonumber\\
&\text{s. to } \check{w}_{ij}=0,~~ \forall i\notin \mathcal{N}_j, ~~~\sum_{i}\check{w}_{ij}=1
\end{align}
where $\check{w}_{ij}$ denotes the $(i,j)$-th entry of $\check{\bbW}$. Upon obtaining $\bbW$ as the constrained least-squares solution of \eqref{eq:lle_kx}, LLE finds $\{\bbpsi_i \}$ that best preserve the neighborhood relationships encoded in $\bbW$ also in the lower dimensional space, by solving
\begin{align}
\label{eq:lle}
&\min_{\bbPsi}\|\bbPsi-\bbPsi{\bbW}\|_F^2\nonumber\\
&~ \text{s.to }~~~ \bbPsi\bbPsi^\top=\bbLambda_d
\end{align}
which is equivalent to 
\begin{align}
\label{eq:lle2}
&\min_{\bbPsi} {\rm tr}[\bbPsi(\bbI-\bbW)(\bbI-\bbW)^\top\bbPsi^\top]\nonumber\\
&~~~~ \text{s. to }~~\bbPsi\bbPsi^\top=\bbLambda_d.
\end{align}
 Conventional LLE adopts $\bbLambda_d=\bbI$, which is subsumed by the constraint in \eqref{eq:lle}. Nonetheless, the difference is just a scaling of $\{\bbpsi_i\}$ when $\bbLambda_d\neq\bbI$. If the diagonal of $\bbLambda_d$ collects the $d$ smallest eigenvalues of matrix $(\bbI-\bbW)(\bbI-\bbW)^\top$, then \eqref{eq:lle}
is a special case of kernel PCA with [cf. \eqref{eq:dPCA4}]
\begin{align}
\label{eq:Kx}
\bbK_y=[(\bbI-\bbW)(\bbI-\bbW)^\top]^{\dagger}.
\end{align}

Similarly, other popular dimensionality reduction methods such as multidimensional scaling (MDS)~\cite{kruskal1978multidimensional}, Laplacian eigenmaps~\cite{belkin2003laplacian}, and isometric feature mapping (ISOMAP)~\cite{Tenenbaum_isomap} can also be viewed as special cases of kernel PCA, by appropriately selecting $\bbK_y$~\cite{ghodsi2006dimensionality}. Thus, ~\eqref{eq:dPCA4} can be viewed as an encompassing framework for nonlinear dimensionality reduction.  This will be the foundation of the general GRAD methods we develop in Sec.~\ref{sec:graph}.
%

\subsection{PCA on graphs}

In several application settings, structural information implying or being implied by dependencies is available, and can benefit the dimensionality reduction task.
 This knowledge can be encoded in a graph and embodied in $\bbPsi$ via graph regularization. Specifically, suppose there exists a graph $\mathcal{G}$ over which the data is smooth; that is, vectors $\{\bbpsi_i\}$ that correspond to connected nodes of ${\cal G}$ are close to each other in Euclidean distance. With $\bbA$ denoting the adjacency matrix of ${\cal G}$, we have $[\bbA]_{ij} = a_{ij}\neq 0$ if node $i$ is connected with node $j$. The Laplacian of ${\cal G}$ is $\bbL_{\cal G}:=\bbD-\bbA$, where $\bbD$ is a diagonal matrix with entries $[\bbD]_{ii} = d_{ii}=\sum_j a_{ij}$. Now consider 
\begin{align}
\label{eq:lap_reg}
	\text{tr}(\bbPsi\bbL_{\cal G}\bbPsi^\top)=\sum_{i=1}^N\sum_{ j\neq i}^N a_{ij}\|\bbpsi_i-\bbpsi_j\|_2^2
\end{align}
which is a weighted sum of the distances of adjacent $\bbpsi_i$'s on the graph. By minimizing \eqref{eq:lap_reg} over $\bbPsi$, the low-dimensional representations corresponding to adjacent nodes with large edge weights $a_{ij}>0$ will be close to each other. Therefore, minimizing \eqref{eq:lap_reg} promotes the smoothness of $\bbPsi$ over the graph.

Augmenting the PCA cost function with the regularizer in \eqref{eq:lap_reg}, yields the graph-regularized PCA \cite{jin2015low}
\begin{align}
\label{eq:gpca}
	\min_{\bbU_d,\bbPsi}\|\bbY-\bbU_d\bbPsi\|_F^2+\lambda\text{tr}(\bbPsi\bbL_{\cal G}\bbPsi^\top)
\end{align}
where $\lambda>0$ is the regularization parameter. 
Building upon \eqref{eq:gpca}, robust versions of graph-regularized PCA have also been developed in e.g. \cite{shang2012graph,shahid2016fast}. Clearly, \eqref{eq:gpca} accounts only for linear dependencies in the data.

%
%
  
\section{GRAD Nonlinear Dimensionality Reduction}
\label{sec:graph}
Other than data correlations, nonlinear dimensionality reduction schemes are not designed to take into account additional prior information. At the same time, PCA on graphs, while able to incorporate prior information in the form of a graph, assumes that data lie near a linear subspace. 
This section will present a novel approach to graph-adaptive nonlinear dimensionality reduction, that encompasses the aforementioned nonlinear dimensionality reduction schemes, as well as linear PCA on graphs.

\subsection{Kernel PCA on graphs}
\label{ssec:gkpca}
Consider the kernel PCA formulation of \eqref{eq:dPCA3}. As with regular PCA, this formulation can  be readily augmented with a graph regularizer, to arrive at
\begin{align}
\label{eq:kPCA_g_}
&\min_{\bbPsi}\|\bbK_y -\bbPsi^\top\bbPsi\|_F^2 + \gamma\text{tr}(\bbPsi\bbL_{\cal G}\bbPsi^\top)\nonumber\\
& \text{s. to }~~\bbPsi\bbPsi^\top=\bbLambda_d
\end{align}
where $\gamma$ is a positive scalar, and $\bbLambda_d$ a diagonal matrix. Since the latter only influences the scaling of $\bbPsi$, for brevity we will henceforth set $\bbLambda_d=\bbI_d$. As kernel PCA can be written as a trace minimization problem [cf.~\eqref{eq:dPCA5}], \eqref{eq:kPCA_g_} reduces to
\begin{align}
\label{eq:kpca_reg}
	&\min_{\bbPsi}-\text{tr}(\bbPsi\bbK_y\bbPsi^\top)+\gamma\text{tr}(\bbPsi\bbL_{\cal G}\bbPsi^\top)\nonumber\\
	& \text{s. to }~~\bbPsi\bbPsi^\top=\bbI.
\end{align}
Combining the Laplacian regularization with the kernel PCA formulation, \eqref{eq:kpca_reg} is capable of finding $\{\bbpsi_i\}$ that preserve the ``lifted'' covariance captured by $\bbK_y$, while at the same time, promoting the smoothness of the low-dimensional representations over the graph $\mathcal{G}$. As a result of the Courant-Fisher characterization~\cite{saad1992numerical},  and with $\bar{\bbK} = \bbK_y-\gamma\bbL_{\cal G}=\bar{\bbV}\bbLambda \bar{\bbV}^\top$, \eqref{eq:kpca_reg} admits a closed-form solution as  
$
	\bbPsi=\bar{\bbV}_d^\top
$,
which denotes the sub-matrix of $\bar{\bbV}$ formed by columns corresponding to the  $d$ largest eigenvalues. When $\gamma$ is set to $0$, one readily obtains the solution of kernel PCA [cf. \eqref{eq:dPCA2}]. 

In addition, instead of directly using $\bbL_{\cal G}$, a family of graph kernels $r^\dag(\bbL_{\cal G}):=\bbU_{\cal G} r^\dag(\bbLambda)\bbU_{\cal G}^\top$ can be employed. Here $r(.)$ is a non-decreasing scalar function of the eigenvalues of $\bbL_{\cal G}$, while $\bbU_{\cal G}$ contains the eigenvectors of $\bbL_{\cal G}$. Introducing $r^\dag(\bbL)$ as a kernel matrix, we have 
\begin{align}
\label{eq:kpca_reg2}
&\min_{\bbPsi}-\text{tr}(\bbPsi\bbK_y\bbPsi^\top)-\gamma\text{tr}(\bbPsi r^\dag(\bbL_{\cal G})\bbPsi^\top)\nonumber\\
& \text{s. to }~~\bbPsi\bbPsi^\top=\bbI.
\end{align}

 By appropriately selecting $r(.)$, different graph properties can be accounted for. As an example, when $r$ sets eigenvalues above a certain threshold to $0$, it acts as a sort of ``low pass'' filter over the graph. Examples of graph kernels are provided in Table~\ref{tab:spectralweightfuns}; see also \cite{romero2017kernel,romero2017kernel-kalman}  on graph kernel options, and the graph properties they capture. 

	\begin{table}[htb]
		\begin{center}
			\begin{tabular}{|p{2.7cm} | p{3.2cm} | p{1.6cm}|}
				\hline
				\textbf{Kernel type}  & \textbf{Function}  & \textbf{Parameters} \\
				\hline
				\hline
				Diffusion ~\cite{kondor2002diffusion}     &
				$r(\lambda)=\exp\{\sigma^2\lambda/2\}$  & 
				$\sigma^2\geq 0$
				\\
				\hline
				$p$-step random walk~\cite{smola2003kernels}    & $r(\lambda) =
				(a-\lambda)^{-p}$ & $a\geq 2$, $p$  \\
				\hline
				Regularized
				Laplacian\cite{smola2003kernels,shuman2013emerging}
				& $r(\lambda)=1 + \sigma^2\lambda$  & $\sigma^2\ge0$ \\
				\hline
				Bandlimited~\cite{romero2017kernel}    &  {\small
				$\begin{aligned}
				\label{eq:defrbl}
				{r}({\lambda}_{{n}}) =  
				\begin{cases}
				1/\beta & n\leq B\\
				\beta & \text{o.w.}
				\end{cases}
				\end{aligned}$}
				&
				$\beta,B>0$
				\\
				\hline
			\end{tabular}
		\end{center}
		\caption{Examples of graph Laplacian kernels.}
		\label{tab:spectralweightfuns}
	\end{table}

As with kernel PCA [cf.~\eqref{eq:dPCA3}] the performance of this approach relies critically on the choice of $\bbK_y$. To circumvent this limitation, the following subsection introduces a multi-kernel based approach to GRAD dimensionality reduction.

\begin{algorithm}[t] 
	\caption{Kernel PCA on graphs }\label{algo:gkpca}
	\begin{algorithmic} 
		\State\textbf{Input:}~ $\bbK_y$, $\bbL_{\cal G}$, $\gamma$, $d$~~~~~~
		
		
		\State {\bf S1.} Find $r(\bbL_{\cal G})=\bb$

		\State{\bf S2.}Find the $d$ largest eigenvalues and corresponding eigenvectors  of 
		$\bbK_y-\gamma r(\bbL_{\cal G})$ and collect them in $\mathbf{V}_d$.
		

		\State {\bf S2.} Find low-dimensional representations $\bbPsi=\bbV_d^\top$.
		
	\end{algorithmic}
\end{algorithm}
\subsection{Multi-kernel learning based approach }
In several application domains, the appropriate kernel for the dimensionality reduction task might be not known a priori. In such cases, one can resort to multi-kernel approaches. Multi-kernel methods select the appropriate kernel function as a linear combination of a number of preselected kernels~\cite{bach2004multiple}. Specifically, $\bbK_y$ can be formed as a linear combination of $Q$ kernel matrices as
\begin{align}
\label{eq:mkl}
\bbK_y=\sum_{q=1}^Q\theta_q\bbK_y^{(q)}
\end{align}
where $\{\bbK_y^{(q)}\}_{q=1}^{Q}$ are predetermined kernel matrices, and $\{\theta_q\}_{q=1}^{Q}$ are unknown non-negative combination weights. Since $\theta_q$'s are non-negative, the resulting $\bbK_y$ is also a valid kernel matrix. Multi-kernel methods ``learn'' the best kernel from the data, by optimizing over the combination weights $\{\theta_q \}_{q=1}^{Q}$.
 Incorporating \eqref{eq:mkl} into \eqref{eq:kpca_reg}, the pertinent optimization problem becomes 
\begin{align}
\label{eq:mkl2}
     \min_{\bm{\theta},\bbPsi}~~ &-\text{tr}(\bbPsi(\sum_{q=1}^Q\theta_q\bbK_y^{(q)})\bbPsi^\top)-\gamma\text{tr}(\bbPsi r^\dag(\bbL_{\cal G})\bbPsi^\top)\nonumber\\
     &\text{s.t.} ~~\bbPsi\bbPsi^\top=\bbI_d\nonumber\\
     & \hspace{5mm}\|\bbtheta\|_2^2\leq 1,~~~\bbtheta\geq\mathbf{0} 
\end{align}
where $\bm{\theta}: = [\theta_1,\ldots,\theta_Q]^{\top}$, and the $\ell_2$-norm regularization is introduced to control the model complexity. As \eqref{eq:mkl2} is non-convex, it will be solved using alternating optimization. When $\{\theta_q\}$ are fixed, \eqref{eq:mkl2} can solved in closed form by eigenvalue decomposition of matrix $\sum_{q=1}^Q\theta_q\bbK_y^{(q)}+\gamma r^\dag(\bbL_{\cal G})$, as in \eqref{eq:kpca_reg2}. With $\bbPsi$ fixed, $\bbtheta$ is found as (see Appendix B for the proof)
\begin{align}
\label{eq:theta}
\theta_q=\frac{\text{tr}(\bbPsi \bbK_y^{(q)}\bbPsi^\top)}{\sqrt{\sum_{q=1}^Q (\text{tr}(\bbPsi \bbK_y^{(q)}\bbPsi^\top))^2}}, ~~q=1\dots, Q.
\end{align}

The overall GRAD multi-kernel(MK)-PCA scheme is tabulated in Algorithm~\ref{algo:gmkpca}.

\begin{algorithm}[t] 
	\caption{Multi-Kernel PCA on graphs }\label{algo:gmkpca}
	\begin{algorithmic} 
		\State\textbf{Input:}~ $\{\bbK_y^{q}\}_{q=1}^{Q}$, $\bbL_{\cal G}$, $\gamma, d$~~~~~~
		
		\While {not converged}
		\State {\bf S1.} Let $\bbK_y = \sum_{q=1}^{Q}\theta_q\bbK_y^{(q)}$
		\State {\bf S2.} Find $\bbPsi$ via Algorithm~\ref{algo:gkpca}
		\State {\bf S3.} Update $\bm{\theta}$ using \eqref{eq:theta}.
		\EndWhile
		
		
		
	\end{algorithmic}
\end{algorithm}

\begin{table*}[htb]
		\begin{center}
			\begin{tabular}{|c | c |c|c|c|}
				\hline
				\textbf{Method}  & \textbf{Formulation} & \textbf{
				Graphs}  & \textbf{Factors} & \textbf{Kernels}\\
				\hline
				\hline
				PCA \cite{jolliffe2002principal}     &
				$\min_{\bbU, \bbPsi}\|\bbY-\bbU\bbPsi \|_F^2$ & No & Yes & No
				\\
				\hline
				GLPCA \cite{jiang2013graph}   & $\min_{\bbU,\bbPsi}\|\bbY-\bbU\bbPsi\|_F^2+\lambda\text{tr}(\bbPsi \bbL\bbPsi^\top)$  &Single&Yes& No\\
				\hline
				LapEmb\cite{belkin2003laplacian} & $\min_{\bbPsi}\text{tr}(\bbPsi \bbL\bbPsi^\top)$& Single & Yes & No\\
				\hline
				RPCAG \cite{shang2012graph} &$\min_{\bbZ,\bbS}\|\bbZ\|_*+\gamma\|\bbS\|_1+\lambda\text{tr}(\bbZ \bbL\bbZ^\top)$ & Single & No & No\\
				\hline
				FRPCAG\cite{shahid2016fast} & \small{$\min_{\bbZ,\bbS}\|\bbY-\bbZ\|_1+\gamma\|\bbS\|_1+\lambda_1\text{tr}(\bbZ \bbL_1\bbZ^\top) +\lambda_2\text{tr}(\bbZ^\top \bbL_1\bbZ)$}&  Two & No& No\\
				\hline
				GRAD KPCA &$\min_{\bm{\theta},\bbPsi,\bbbeta} -\text{tr}(\bbPsi(\sum_{q=1}^Q\theta_q\bbK_y^{(q)})\bbPsi^\top)-\gamma\text{tr}(\bbPsi (\sum_{m=1}^M \beta_mr^{-1}(\bbL_{\cal G}^m))\bbPsi^\top)$ & Multiple& Yes& Multiple\\
				\hline
			\end{tabular}
		\end{center}
		\caption{Comparison of graph-regularized PCA methods.} 
		\label{tab:competingalgos}
	\end{table*}


	Even though only a single graph regularizer is introduced in \eqref{eq:kpca_reg}, our method is flexible to include \emph{multiple} graph regularizers based on different graphs. Therefore, the proposed method offers a powerful tool for dimensionality reduction with prior information encoded in the so-called \emph{multi-layer} graphs~\cite{traganitis2017,kivela2014multilayer}.
	Suppose that $L$ such $N$-node graphs $\{\mathcal{G}_\ell\}_{\ell=1}^{L}$ are available, each with corresponding  Laplacian matrices $\{\bbL_{\cal G}^{\ell} \}_{\ell=1}^{L}$. If all $L$ graphs are expected to have the same contribution then multiple Laplacian regularizers, say one per graph, can be introduced in the objective function of \eqref{eq:mkl2} as
	\begin{align}
	\label{eq:mkl3}
	\min_{\bm{\theta},\bbPsi}~~ &-\text{tr}(\bbPsi(\sum_{q=1}^Q\theta_q\bbK_y^{(q)}\bbPsi^\top)-\gamma\sum_{
	\ell=1}^{L}\text{tr}(\bbPsi r^{\dag}(\bbL_{\cal G}^{\ell})\bbPsi^\top)\nonumber\\
	&\text{s.t.} ~~\bbPsi\bbPsi^\top=\bbI_d\nonumber\\
	& \hspace{5mm}\|\bbtheta\|_2^2\leq 1,~~~\bbtheta\geq\mathbf{0}.
	\end{align}
When the appropriate graph regularizer is unknown, a scheme similar to the multi-kernel approach of \eqref{eq:mkl2} can be employed to choose the appropriate graph kernel. In this case, the graph kernel can be expressed as a linear combination of the $M$ available graph regularizers, that is
	$r^{\dag}(\bbL_{\cal G}) = \sum_{\ell=1}^{L}\beta_{\ell} r^{\dag}(\bbL_{\cal G}^{\ell})$, where $\beta_{\ell}$ are unknown non-negative combination weights.
	Introducing this multi-graph kernel term into \eqref{eq:mkl2} yields
\begin{align} 
\label{eq:mmkpca}
	 \min_{\bm{\theta},\bbPsi,\bbbeta}~~ &-\text{tr}(\bbPsi(\sum_{q=1}^Q\theta_q\bbK_y^{(q)})\bbPsi^\top)-\gamma\text{tr}(\bbPsi (\sum_{\ell=1}^L \beta_{\ell}r^{\dag}(\bbL_{\cal G}^{\ell}))\bbPsi^\top)\nonumber\\
     &\text{s.t.} ~~\bbPsi\bbPsi^\top=\bbI_d\nonumber\\
     & \hspace{5mm}\|\bbtheta\|_2^2\leq 1,~~~\bbtheta\geq\mathbf{0}\nonumber \\
     & \hspace{5mm}\|\bbbeta\|_2^2\leq 1,~~~\bbbeta\geq\mathbf{0}
\end{align}
where $\bbbeta:=[\beta_1, \dots, \beta_L]^\top$.
Similar to \eqref{eq:mkl2},  the non-convex problem in \eqref{eq:mmkpca} can be solved in an alternating fashion. With $\bbbeta$ fixed, $\bbPsi$ can be found in closed form by eigenvalue decomposition of matrix $	\sum_{q=1}^Q\theta_q\bbK_y^{(q)}+\gamma \sum_{\ell=1}^L \beta_{\ell}r^{\dag}(\bbL_{\cal G}^{\ell})$, while $\bbtheta$ can be obtained using \eqref{eq:theta}. When $\bbPsi$ and $\bbtheta$ are fixed, $\bbbeta$ can be found in closed form as
\begin{align}
	\beta_{\ell}=\frac{\text{tr}(\bbPsi r^{\dag}(\bbL_{\cal G}^{\ell}))\bbPsi^\top)}{\sqrt{\sum_{\ell=1}^L (\text{tr}(\bbPsi r^{\dag}(\bbL_{\cal G}^{\ell}))\bbPsi^\top))^2}},~~~ \ell=1, \dots,L.
\end{align}
%

In this section, we put forth a novel scheme for dimensionality reduction over graphs that can also capture nonlinear data dependencies; see also Table \ref{tab:competingalgos} that  summarizes how the novel approach fits within the context of prior relevant works. This table showcases the optimization problem solved by each algorithm, as well as whether prior information in the form of a graph can be incorporated. In addition, Table~\ref{tab:competingalgos} indicates if low-dimensional representations are directly provided by an algorithm, and whether kernels have been employed to capture data correlations.

\section{Generalizations}
\label{sec:general}
The present section showcases three generalizations and applications of our novel GRAD dimensionality reduction scheme. Specifically, the following subsections extend the methods of Sec.~\ref{sec:graph} to multi-modal datasets, and semi-supervised settings, as well as generalize the LLE.

\subsection{Dimensionality reduction for multi-modal datasets}
As discussed in Sec.~\ref{sec:intro}, many datasets comprise multi-modal data, that is  data with features belonging to different types, such as binary, categorical or real-valued features.
  In this subsection, we demonstrate how our proposed GRAD nonlinear dimensionality reduction approach can readily handle such cases. 

Suppose that the collected $N$ data contain $M$ different modes. Vectors of mode $m$ have dimension $D_m$, and are collected in a $D_m\times N$ submatrix $\bbY_m$. With these $M$ sets of vectors at hand, $M$ different graphs $\{\mathcal{G}_m\}_{m=1}^{M}$, each with $N$ nodes can be inferred, based on possibly diverse similarity metrics. These metrics can be different for each mode, e.g. graphs for binary data can be constructed based on the Hamming distance, while graphs for real-valued data can be based on linear or nonlinear correlations. These $M$ graphs can be considered as an $M$-layer multiplex graph~\cite{kivela2014multilayer}, on which our proposed scheme can be readily applied. Specifically, given the Laplacian matrices for each of these $M$ graphs $\{\bbL_{\cal G}^1, \cdots, \bbL_{\cal G}^M\}$, lower dimensional representations can be obtained as

%
\begin{align}
\label{prob:mldr}
	&\min_{\bbPsi}-\sum_{m=1}^M\text{tr}(\bbPsi r^\dag(\bbL_{\cal G}^m)\bbPsi^\top)\nonumber\\
	& \text{s. to }~~\bbPsi\bbPsi^\top=\bbI_d.
\end{align}
%
Therefore, the complexity of GRAD dimensionality reduction is in the order of $\mathcal{O}(DN^2)$, which is the same as the dual PCA. However, the graph-based PCA now can handle data that consist of heterogeneous features, e.g. binary, categorical or real-valued.

This scheme can also be used for dimensionality reduction of very high-dimensional data ($D\gg$). The $D\times N$ data matrix $\bbY$ can be split,  into $M$ submatrices $\{\bbY_m \}_{m=1}^{M}$ each of dimension $D_m\times N$. These submatrices may contain non-overlapping or overlapping subsets of each data vector $\{\bm{y}_i\}_{i=1}^{N}$. Creating a graph for each $\bbY_m$, \eqref{prob:mldr} can be used to find lower dimensional representations $\bbPsi$.

\subsection{Semi-supervised dimensionality reduction over graphs}
\label{ssec:semi}
In this subsection, we develop our proposed scheme for semi-supervised dimensionality reduction. In addition to data samples $\{\bby_i \}_{i=1}^{N}$,  domain knowledge here becomes available in the form of a few pairwise
constraints. These constraints specify whether a pair of data samples belong to the same class (must-link constraints), or to different classes (cannot-link constraints). Specifically, let $\mathcal{S}$ be the set containing the tuples $(i,j)$ for some data belonging to the same class (must-link constraints), and $\mathcal{D}$  the set containing the tuples corresponding to data from different classes (cannot-link constraints). Given these two sets, two graphs can be constructed, one for each constraint set. The graph $\mathcal{G}^{\mathcal{S}}$ is constructed based on the must-link constraints with adjacency matrix $\bbA^{\mathcal{S}}$ having entries
 \begin{align}
 \label{graph:s}
 [\bbA^{\mathcal{S}}]_{ij} = \begin{cases}
 1, \quad \text{ if } (i,j)\in\mathcal{S} \\
 0, \quad \text{ otherwise.}
 \end{cases}
 \end{align}
 Similarly, the graph $\mathcal{G}^{\mathcal{D}}$ is constructed based on the cannot-link constraints 
 and its adjacency matrix $\bbA^{\mathcal{D}}$ has entries
 \begin{align}
 \label{graph:d}
 [\bbA^{\mathcal{D}}]_{ij} = \begin{cases}
 1, \quad \text{ if } (i,j)\in\mathcal{D} \\
 0, \quad \text{ otherwise.}
 \end{cases}
 \end{align}
 Letting $\bbL_{\mathcal{G}}^{\mathcal{S}}$, $\bbL_{\mathcal{G}}^{\mathcal{D}}$ denote the graph Laplacians of $\mathcal{G}^{\mathcal{S}}$ and $\mathcal{G}^{\mathcal{D}}$ respectively, the low-dimensional representations of $\bbY$ can be obtained as follows
\begin{align}
\label{eq:kpca_semi}
&\min_{\bbPsi}-\text{tr}(\bbPsi\bbK_y\bbPsi^\top)+\gamma_1\text{tr}(\bbPsi\bbL^{\cal S}_{\cal G}\bbPsi^\top)-\gamma_2\text{tr}(\bbPsi\bbL^{\cal D}_{\cal G}\bbPsi^\top)\nonumber\\
&\text{s.t.} \bbPsi\bbPsi^\top=\bbI
\end{align}
where $\gamma_1,\gamma_2 >0$ are regularization constants.

Clearly, the term $\text{tr}(\bbPsi\bbL^{\cal S}_{\cal G}\bbPsi^\top)$ forces the low-dimensional representations corresponding to the must-link constraints to be close, while the term 
$-\text{tr}(\bbPsi\bbL^{\cal D}_{\cal G}\bbPsi^\top)$ ``pushes'' data corresponding to the cannot-link constraints away from each other. The GRAD regularizers effecting these two constraints are well motivated when one is interested in classifying high-dimensional vectors. If only a few of these vectors are labeled, such a semi-supervised setting should be accounted for in obtaining the low-dimensional representations based on which classification is to be performed subsequently.

\begin{figure*}[t]
	\begin{minipage}[b]{.49\textwidth}
		\centering
		\includegraphics[width=9.5cm]{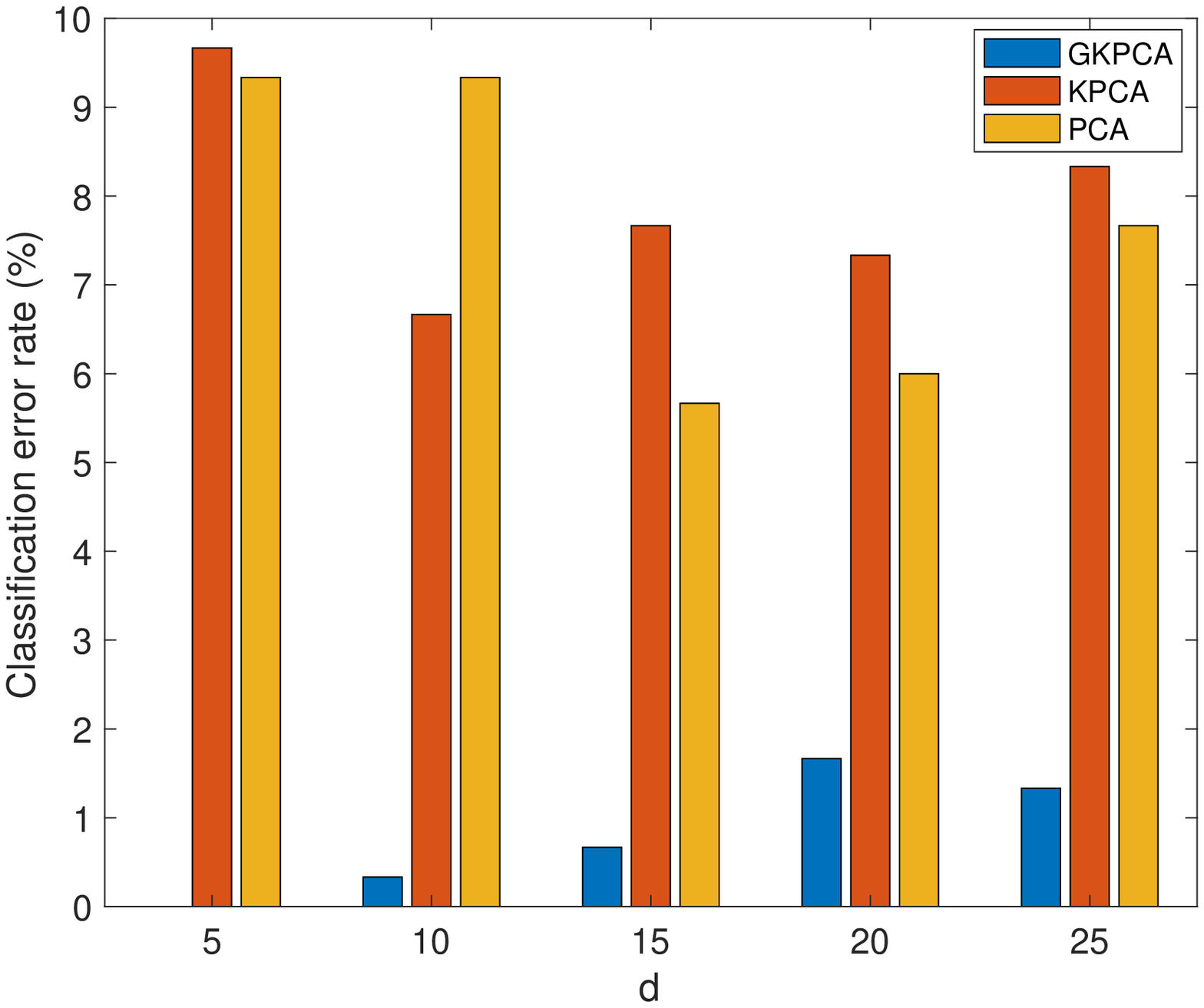}
		\centerline{(a)}
	\end{minipage}
	\begin{minipage}[b]{.49\textwidth}
		\centering
		\includegraphics[width=9.5cm]{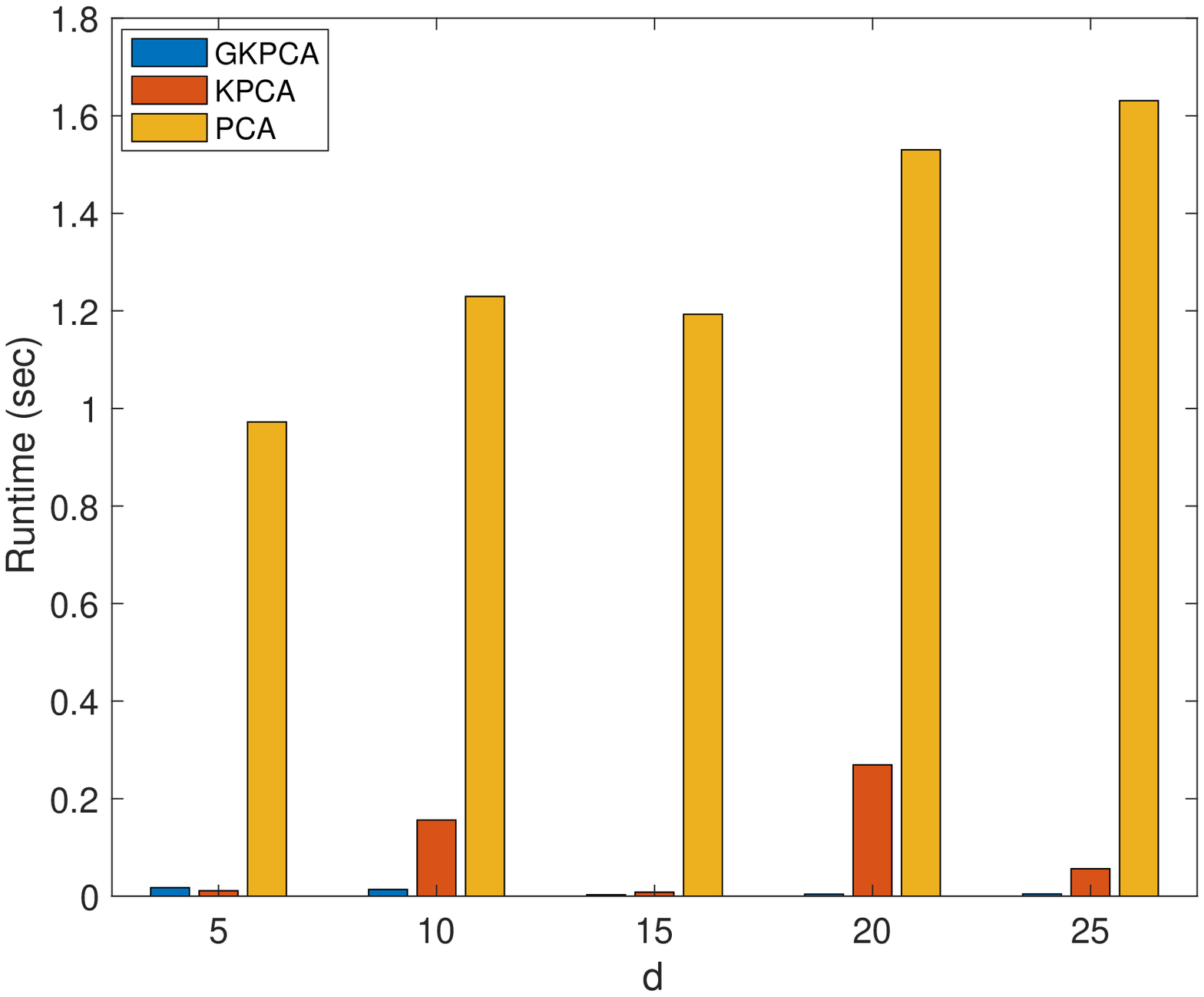}
		\centerline{(b)}
	\end{minipage}
	\vspace{2mm}
	\caption{{Classification based on $\{\bbpsi_i\}_{i=1}^N$ assessed by: (a) Classification error rate; and, (b) Running time;  } }
	\label{fig:face}
\end{figure*}

\begin{figure*}[t]
	\centering
	\includegraphics[width=19cm]{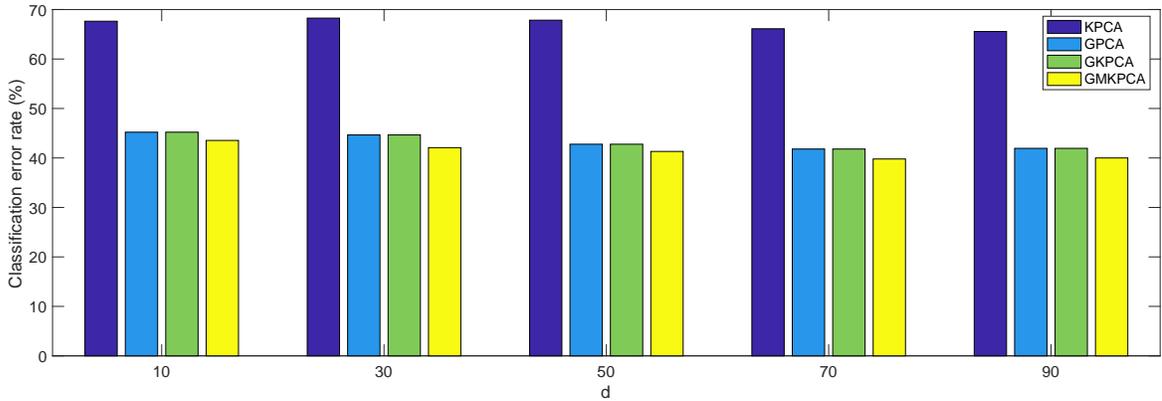}
	\vspace{2mm}
	\caption{Clustering results using $\{\bbpsi_i\}_{i=1}^N$ on COIL20 dataset.  }
	\label{fig:cluster}
\end{figure*}

\begin{algorithm}[t] 
	\caption{Local nonlinear embedding over graphs }\label{algo:lneg}
	\begin{algorithmic} 
		\State\textbf{Input:}~ $\bbY$, $\bbL_{\cal G}$ $\gamma,d$~~~~~~
		
		
				\State {\bf S1.} Estimate ${\bbW}$ from $\bbY$.
				\State {\bf S2.} Obtain kernel matrix $\bbK_y$ via \eqref{eq:Kx}.


		\State {\bf S3.} Find $\bbPsi$ as the leading eigenvectors of $\bbK_y$.
		
	\end{algorithmic}
\end{algorithm}

\subsection{Local nonlinear embedding on graphs}
In this subsection, we develop a major GRAD enhancement of the well appreciated nonlinear dimensionality reduction effected by LLE. We refer to our novel scheme as local nonlinear embedding on graphs (LNEG), because it can capture both linear and nonlinear dependencies among neighboring data, in addition to the structure induced by the graph $\mathcal{G}$. To this end, suppose that each data vector can be represented by its neighbors entry-wise as 
 \begin{align}
 	[\bby_i]_m=\sum_{j\in \mathcal{N}_i}h_{ij}\big([\bby_j]_m\big)+[\bbe_i]_m, ~~ m=1,\dots, D
	\label{eq:sem:nonlinear}
 \end{align}
 where $\{h_{ij}(\cdot)\}_{i,j=1}^N$ are prescribed scalar nonlinear functions admitting a $P$th-order expansion
 \begin{align}
 	h_{ij}(z)=\sum_{p=1}^Pw_{ij}[p]z^p
\label{eq:h}
 \end{align}
and coefficients $\{w_{ij}[p]\}$ are to be determined. Taylor's expansion asserts that for $P$ sufficiently large, \eqref{eq:h} offers an accurate approximation for all memoryless differentiable nonlinear functions. Such a nonlinear model has been used for graph topology identification~\cite{shen2017tsp}, but we here employ it as a first step of our LNEG scheme implementing the local nonlinear embedding. In vector form, \eqref{eq:sem:nonlinear} becomes
 \begin{align}
 \label{eq:vym}
 \bar{{\bby}}_m^\top=\tilde{\bby}_m^\top\tilde{\bbW}+\bbe_m
 \end{align} 
 where $\bar{\bby}_m^\top:=[y_{1m} \dots y_{Nm}]$ denotes the $m$-th row of $\bbY$; the extended vector on the right hand side of \eqref{eq:vym} is $\tilde{\bby}_m^\top{ := }[\tilde{\bby}_{1m}^\top ~\tilde{\bby}_{2m}^\top~\cdots ~\tilde{\bby}_{Nm}^\top]$ formed
 with sub-vectors $\tilde{\bby}_{im}{ := }[y_{im},~y_{im}^2,\cdots,~y_{im}^P]^\top$; and, the $N\times NP$ matrix  $\tilde{\bbW}$ is defined as
 \begin{align}
 \tilde{\bbW}:=\left[
 \begin{array}{cll}
 \bbw_{11} &	\cdots &\bbw_{1N}\\
 \vdots & & \vdots\\
 \bbw_{N1}&\cdots	& \bbw_{NN}\end{array} \right]\label{def:W}
 \end{align}
 where the entries of $\bbw_{ij}:= \left[w_{ij}[1], \dots, w_{ij}[P] \right]^\top$ are the coefficients in \eqref{eq:h}, specifying the nonlinear correlations between data. Upon defining $\tilde{\bbY}:=[\tilde{\bby}_1~\cdots~\tilde{\bby}_D]^\top$, one obtains the following nonlinear matrix model
 \begin{align}
 \bbY=\tilde{\bbY}\tilde{\bbW}+\bbE.
 \label{mod:nonlinear}
 \end{align}
 Matrix $\tilde{\bbW}$ can be estimated using the least-squares (LS) or sparse regularized LS criteria, e.g.,
 \begin{align}
 \label{prob:sems}
 	\bbW^{*}=\arg \min\| \bbY-\tilde{\bbY}\tilde{\bbW}\|_F^2+\|\tilde{\bbW}\|_1
 \end{align}
 which is convex but non-smooth, and can be solved iteratively to attain the global optimum using proximal splitting methods, see e.g. \cite{shen2017tsp} for details.
  Using ${\bbW}^{*}$, an $N\times N$ matrix $\bbW$, similar to the one derived for LLE, can be obtained.
  Different from LLE, where $\bf W$ specifies tangential hyperplanes, our generalization here allows the local geometry to be captured by tangential nonlinear manifolds. Since $h$ can also be linear, LNEG is expected to perform at least as well as the LLE. 
 With the estimated $\bbW$ at hand, the low-dimensional representations can be obtained via \eqref{eq:lle2}; see also  Algorithm \ref{algo:lneg}.

\begin{figure*}[t]
	\begin{minipage}[b]{.49\textwidth}
		\centering
		\includegraphics[width=9.cm]{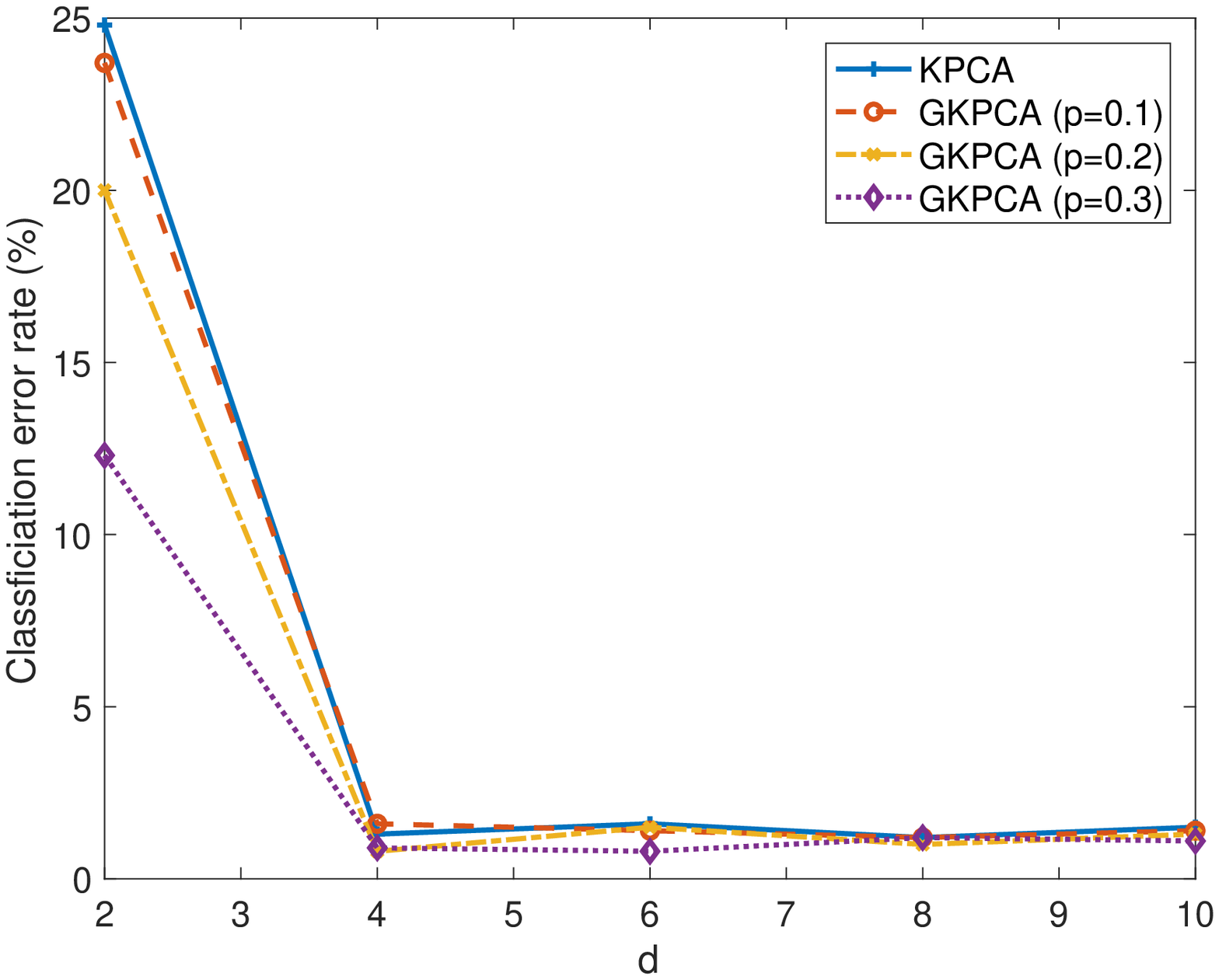}
		\centerline{(a)}
	\end{minipage}
	\begin{minipage}[b]{.49\textwidth}
		\centering
		\includegraphics[width=9.cm]{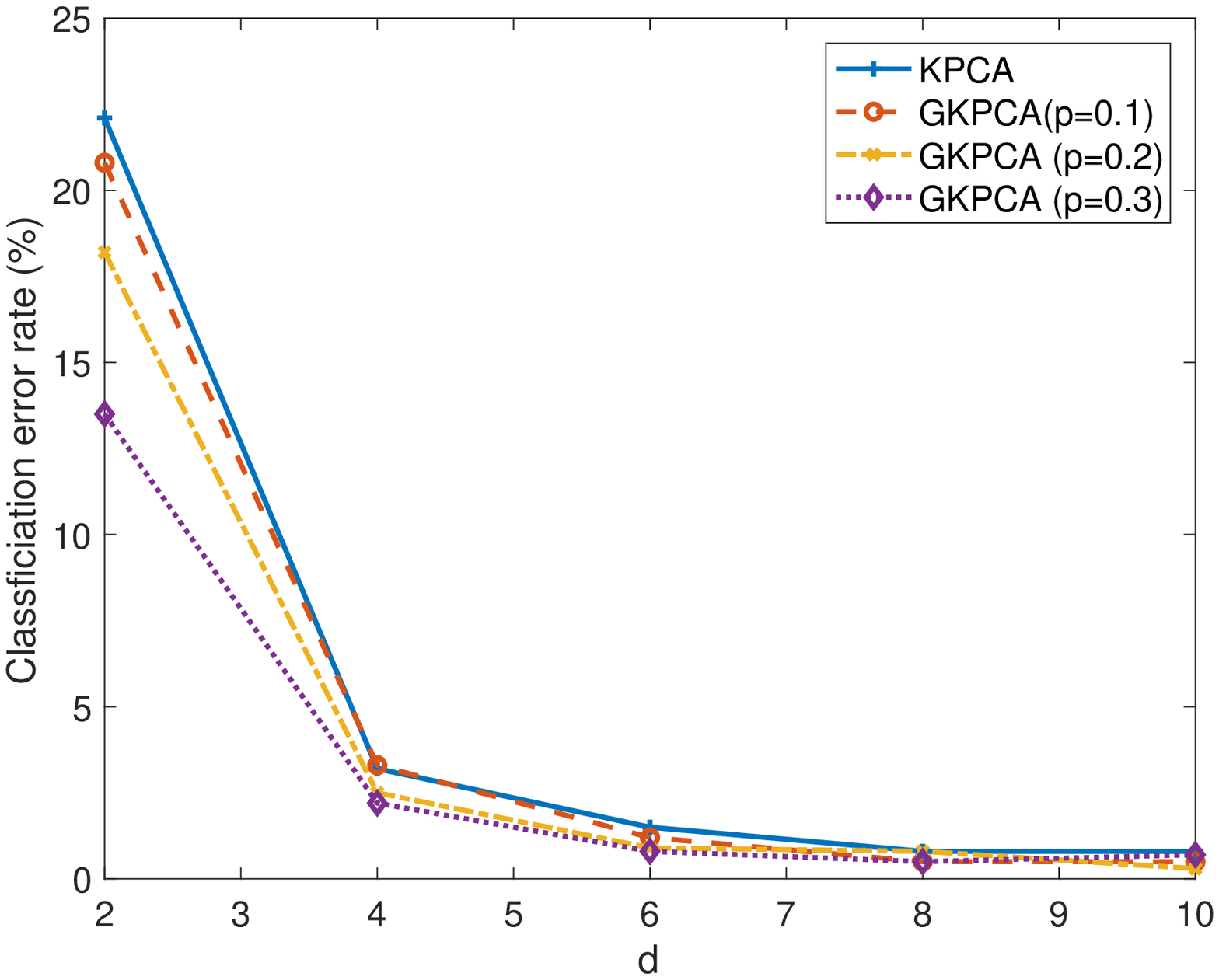}
		\centerline{(b)}
	\end{minipage}
	\vspace{2mm}
	\caption{{Classification for USPS dataset using $\{\bbpsi_i\}_{i=1}^N$ for (a) Digits 5 and 6 (b) Digits 7 and 8.  } }
	\label{fig:semi}
\end{figure*}

\begin{figure*}[t]
	\begin{minipage}[b]{.49\textwidth}
		\centering
		\includegraphics[width=9.5cm]{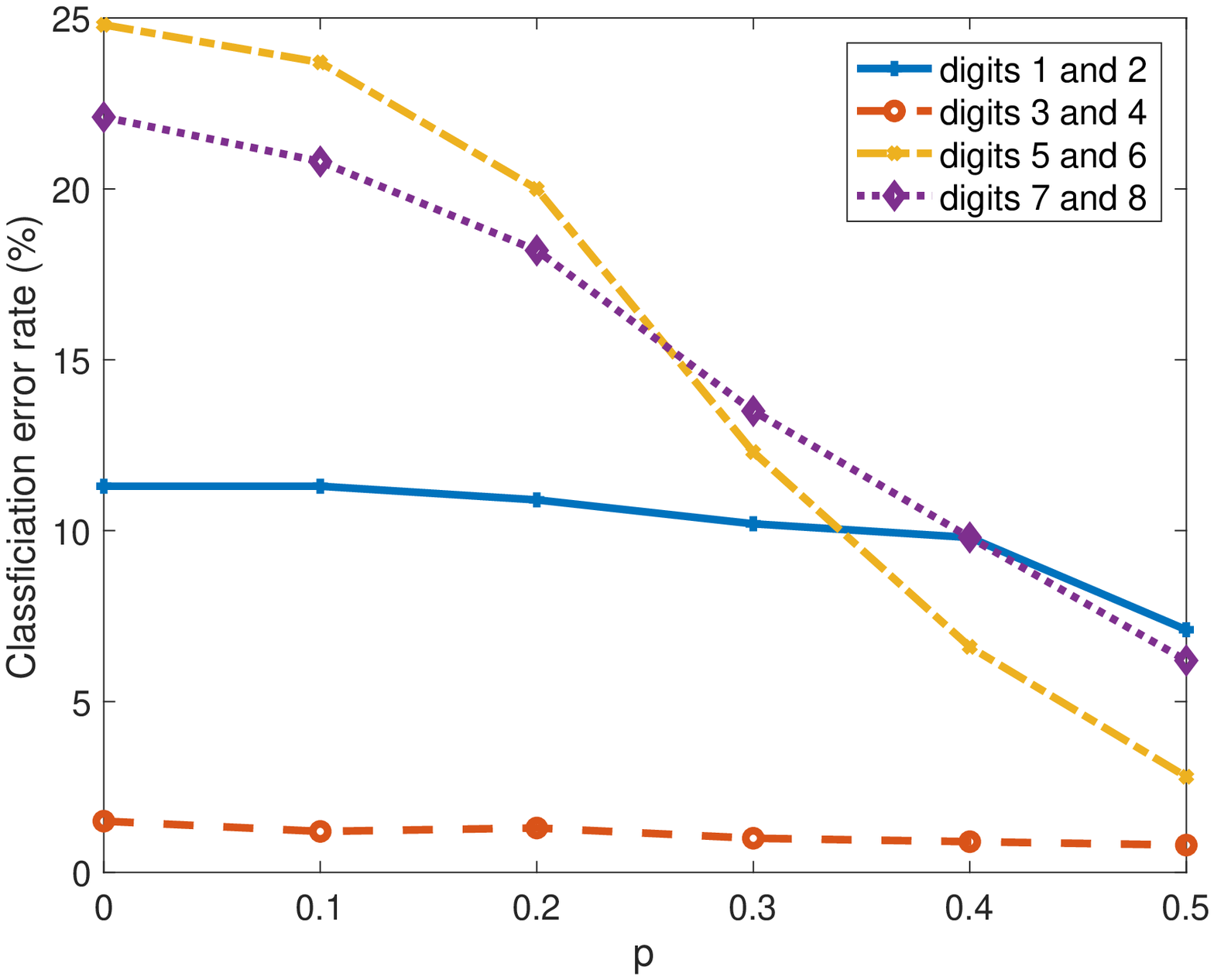}
		\centerline{(a)}
	\end{minipage}
	\begin{minipage}[b]{.49\textwidth}
		\centering
		\includegraphics[width=9.5cm]{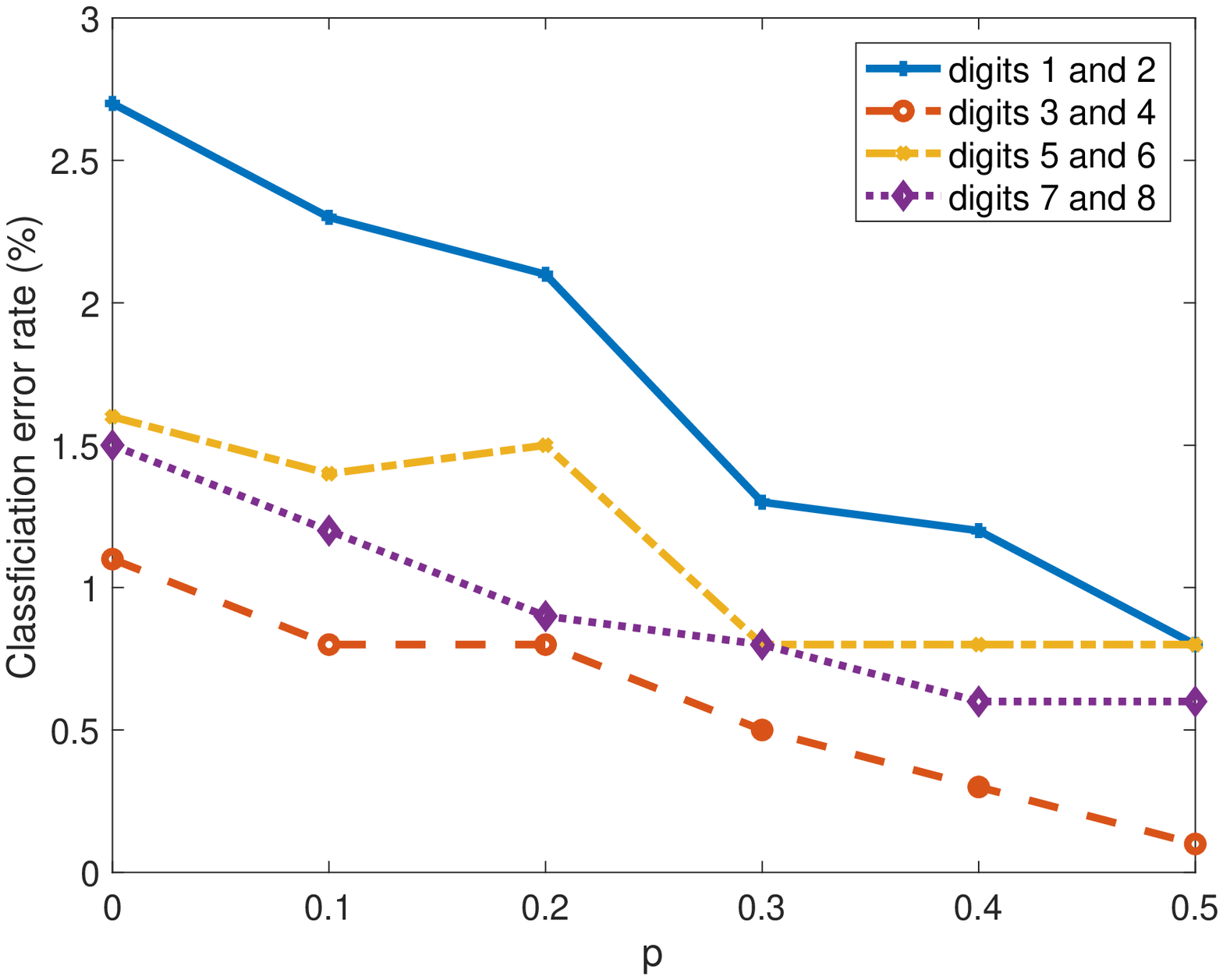}
		\centerline{(b)}
	\end{minipage}
	\vspace{2mm}
	\caption{{Classification for USPS dataset using  $\{\bbpsi_i\}_{i=1}^N$ with: (a) $d=2$; and (b) $d=6$;  } }
	\label{fig:semi2}
\end{figure*}

\section {Numerical Tests}\label{sec:numerical_tests}
The performance of our proposed algorithms, as well as their generalizations are tested in the present section. Numerical tests are carried out on both synthetic and real datasets. The performance of the dimensionality reduction task is evaluated through classification and clustering experiments.
 Specifically, the clustering and classification algorithms used are $K$-means and support vector machines (SVMs), respectively.  The software used to conduct all experiments is MATLAB~\cite{MATLAB:2016}. Reported results represent the averages of $50$ independent Monte Carlo runs.
For both clustering and classification tests, performance is measured using the \emph{error rate}, which is defined as the
percentage of mis-clustered/ misclassified samples:
	\begin{equation*}
	\text{Error Rate} := 1-\frac{\text{\#  data correctly clustered/classified}}{N}\times 100\%.
	\end{equation*}
The datasets used  are the following:  
\begin{itemize}
	\item USPS image dataset \cite{hull1994database}: This consists of $N=9,298$ images of size $16 \times 16$. Each image contains a digit scanned from U.S. Postal Service envelopes, and the dataset consists of $K=10$ classes, one per digit.
	\item Coil20 image dataset \cite{nene1996columbia}: This contains $N=1,440$ $32\times 32$
images of $K=20$ objects. For each object $72$ images are available, each taken under a different angle.
	\item Drivface image dataset \cite{diaz2016reduced}: This consists of $N=66$ images of size $80\times 80$, depicting images of drivers from two different angles, front images or side images.
\end{itemize} 
Properties of these datasets are summarized in Table \ref{tab:dataset}.

\begin{table}[h]
	\begin{center}
		\begin{tabular}{c | c | c|c}
			\hline
			\textbf{Dataset}  & $N$ \textbf{samples}  & \textbf{$D$ features} &$K$ \textbf{classes}\\
			\hline
			\hline
			Driveface& $66$& $6,400$& $2$\\
			\hline
			USPS     &$9,298$ & $256$ & $10$
			\\
			\hline
			COIL20  &  $1,440$ & $1,020$ & $20$  \\
			\hline
			
		\end{tabular}
	\end{center}
	\caption{Datasets description.}
	\label{tab:dataset}
\end{table}



\subsection{Graph Kernel PCA { and Graph Multi-kernel PCA}}
{In this section, the performance of the GRAD MK-PCA and GRAD KPCA algorithms is evaluated using both classification and clustering tests.}

\noindent\textbf{Classification experiment.}
In this experiment, Algorithm~\ref{algo:gkpca} (abbreviated henceforth as \emph{GKPCA}) is tested on the Drivface dataset. The vectorized images $\bby_i\in\mathbb{R}^{6,400}$ are used as columns of $\bbY$. Here GKPCA is compared to PCA and KPCA. 
For the novel GKPCA and the KPCA algorithm, a Gaussian kernel with bandwidth $\sigma^2=1$ is employed. For GKPCA, the graph is constructed by pairwise linear correlation coefficients of feature vectors $\{\bbf_i\}$ extracted from the facial images, each $\bbf_i$ collecting the coordinates of nose, eyes and ears in the picture. Note that, this feature information is provided in the dataset.

{Each dimensionality reduction algorithm is applied on $\{\bby_i\}$ and low dimensional representations $\{\boldsymbol{\psi}_i\}$ are obtained.}
Upon obtaining  $\{\bbpsi_i\}$, classification is performed using a linear SVM  with $5$ fold cross validation, with $80\%$ of the data  used for training and the remaining $20\%$ for testing.


 Figure \ref{fig:face} (a) depicts the testing classification error rate for different values of $d$. Clearly, the novel GKPCA approach outperforms both KPCA and PCA. In addition, Figure \ref{fig:face} (b) shows the runtime of different algorithms and corroborates that the kernel based approaches perform much faster that PCA, because $D\gg N$. It can also be seen that GKPCA is more computationally efficient than Kernel PCA for most values of $d$. This is due to the graph regularization, which makes the $\bar{\bbK}$ [cf.~Sec.~\ref{ssec:gkpca}] matrix well-conditioned, and thus speeds up the eigenvalue decomposition. 
%


%

\noindent\textbf{Clustering experiment.} In this experiment, the clustering performance was tested using $\{\bbpsi_i\}$ obtained from different algorithms. GKPCA and Alg.~\ref{algo:gmkpca} (termed henceforth as \emph{GMKPCA}) are compared to KPCA and GPCA. 
For the GKPCA and KPCA algorithms, a Gaussian kernel with bandwidth $\sigma^2=1$ is employed. For GPCA, GKPCA and GMKPCA, the graph used is constructed by finding the pairwise correlation coefficients, $\bar{a}_{ij}=\frac{\bby_i^\top\bby_j}{\|\bby_i\|_2\|\bby_j\|_2}$, and connecting each data sample with its $100$ neighbors having the largest $\bar{a}_{ij}$; that is, $a_{ij}=\bar{a}_{ij}$ if $j\in\mathcal{N}_i$, otherwise $a_{ij}=0$.
 The GMKPCA uses a dictionary of Gaussian kernels with bandwidths $\sigma^2$ taking $10$ equispaced values from $0.01$ to $1$.
The $K$-means algorithm was repeated $50$ times and the best result was reported. 
{Figure~\ref{fig:cluster} shows the clustering error rates for different algorithms, and after varying $d$ for the Drivface dataset. Clearly, GMKPCA outperforms the alternatives for clustering tasks.}



%

\subsection{Semi-supervised graph-based dimensionality reduction}

In this subsection, the semi-supervised dimensionality reduction scheme of Sec.~\ref{ssec:semi}
is evaluated on the USPS dataset. 
For each experiment, a set of $1,000$ images of two different digits is used, with $500$ images for each digit. After obtaining low dimensional representations, a linear SVM classifier with $5$-fold cross validation was used to distinguish the two digits.

Let $\Omega$ denote the set containing the indices of data for which labels are available. Two graphs were generated using \eqref{graph:s} and \eqref{graph:d} based on the known labels. Figure \ref{fig:semi} showcases the performance of Algorithm \ref{algo:gkpca} as a function of $d$ and for different numbers of available labels $|\Omega|=p\times N$, { when classifying the digits $5$ and $6$ or $7$ and $8$}. The available labeled data are chosen uniformly at random for each experiment.  As the number of known labels increases, the classification error rate also decreases. Figure \ref{fig:semi2} depicts the classification error rate for a variable number of available labels used to classify different digits with $d=2$ and $d=6$, respectively. During all experiments, $\gamma_1$ and $\gamma_2$ were both set at $0.5$, and a Gaussian kernel with $\sigma^2=1$ was used.  Clearly, this semi-supervised scheme can successfully incorporate label information into the graph-adaptive nonlinear dimensionality reduction task, such that the ensuing classification performance is improved.


%
\begin{figure}[t]
	\vspace{1mm}
	\begin{minipage}[b]{.24\textwidth}
		\centering
		\includegraphics[width=4.8cm]{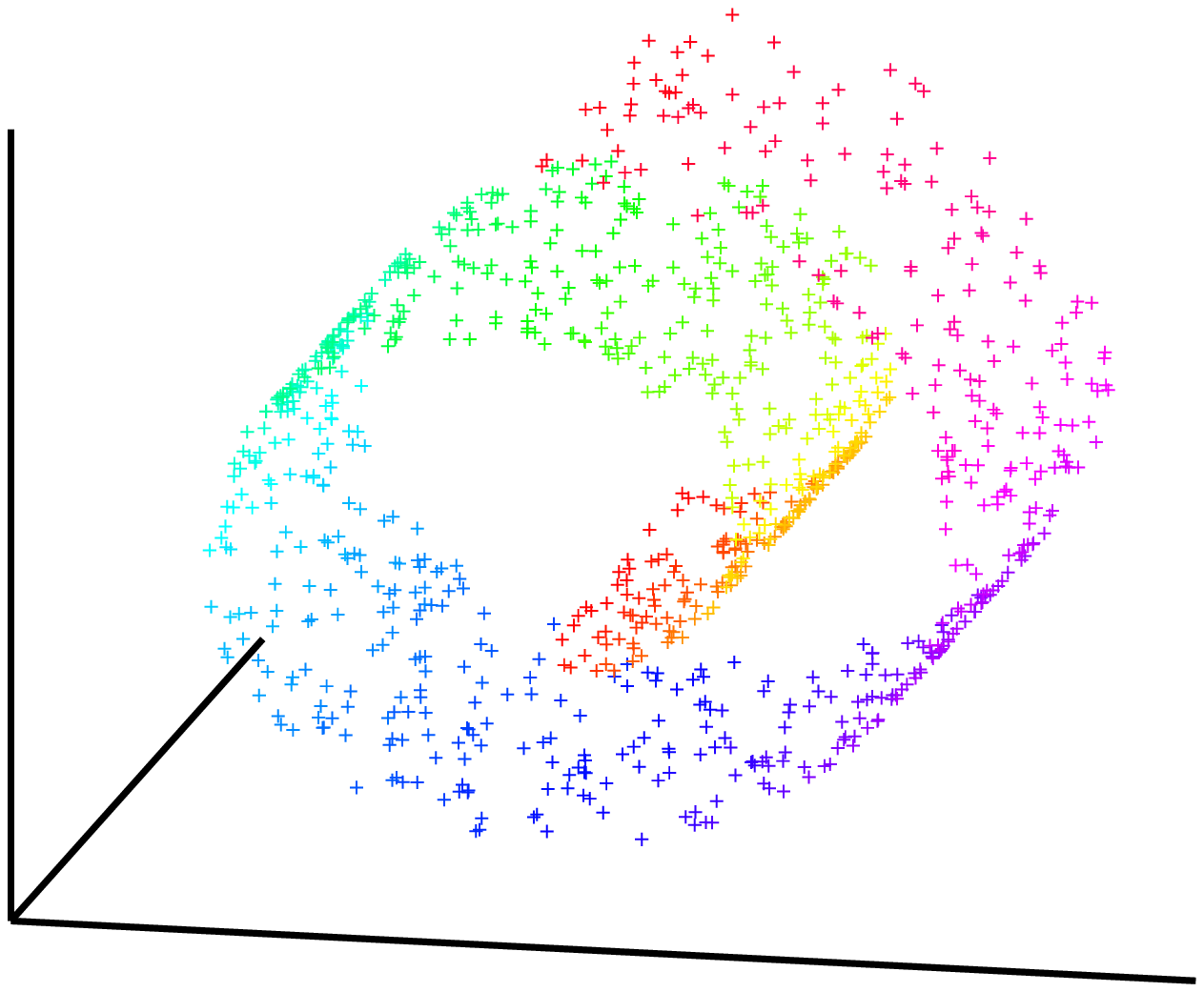}
		\centerline{(a)}
	\end{minipage}
	\begin{minipage}[b]{.24\textwidth}
		\centering
		\includegraphics[width=4.8cm]{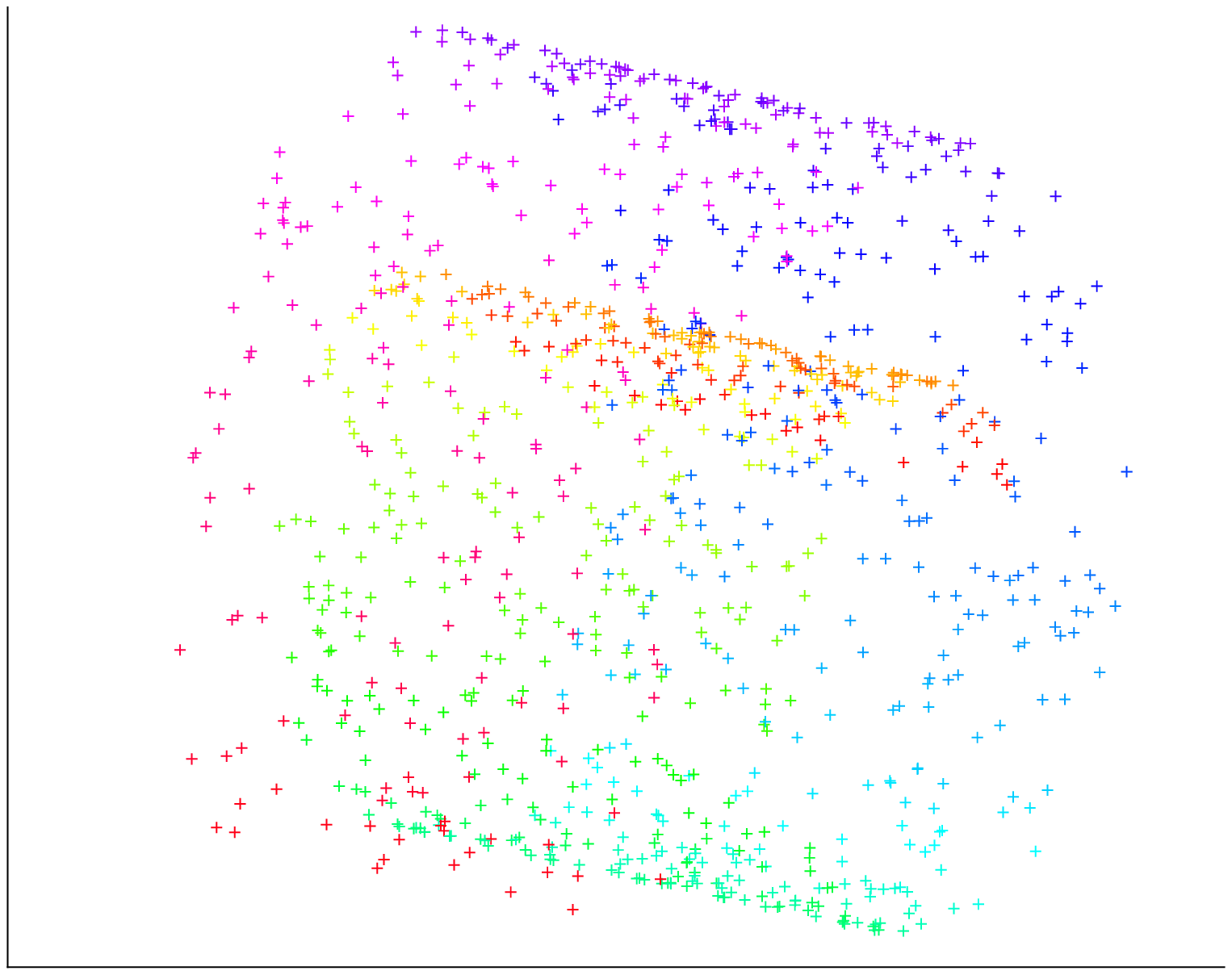}
		\centerline{(b)}
	\end{minipage}
	%
	%
	\begin{minipage}[b]{.24\textwidth}
		\centering
		\includegraphics[width=4.8cm]{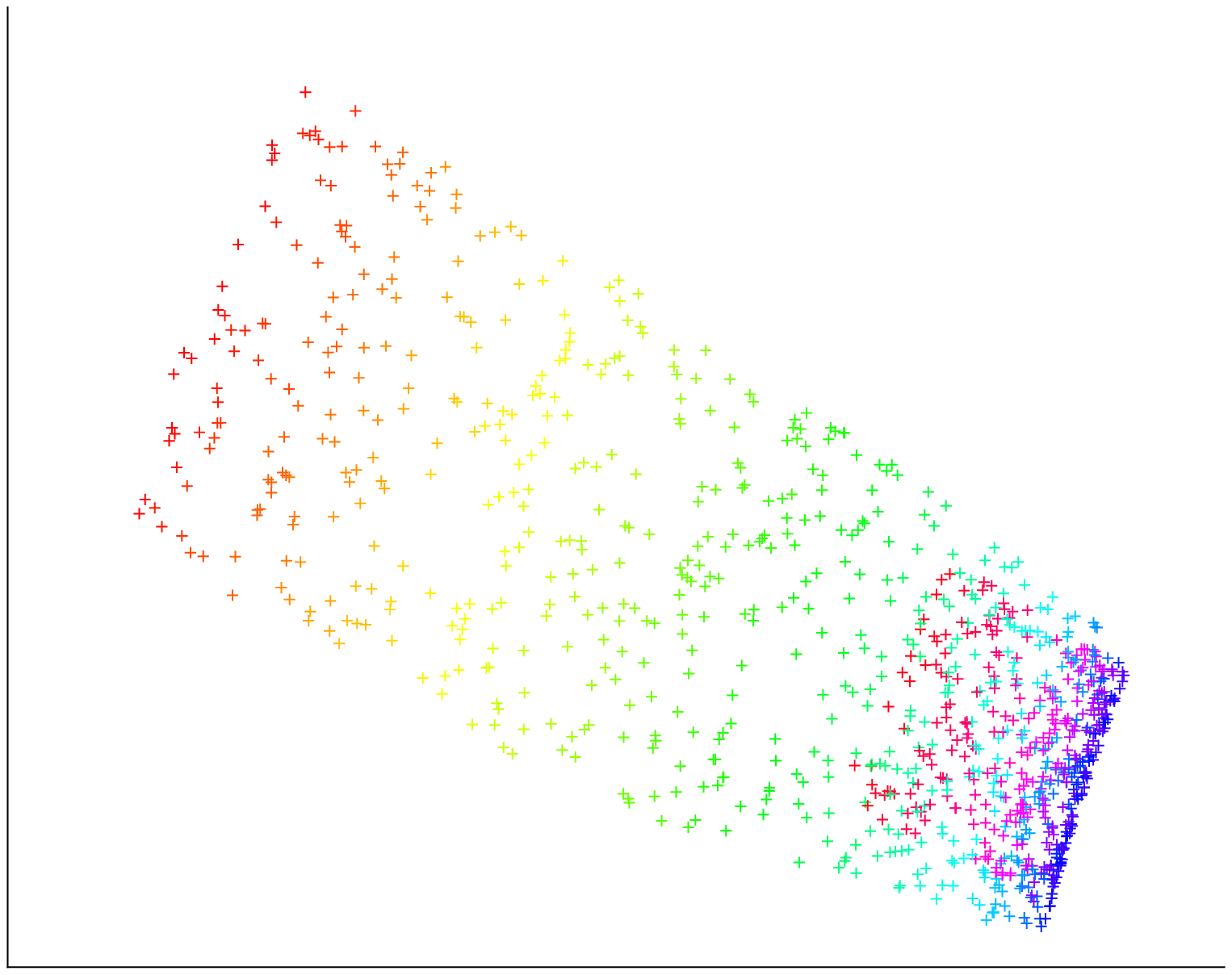}
		\centerline{(c)}
	\end{minipage}
	\vspace{1mm}
	\begin{minipage}[b]{.24\textwidth}
		\centering
		\includegraphics[width=4.8cm]{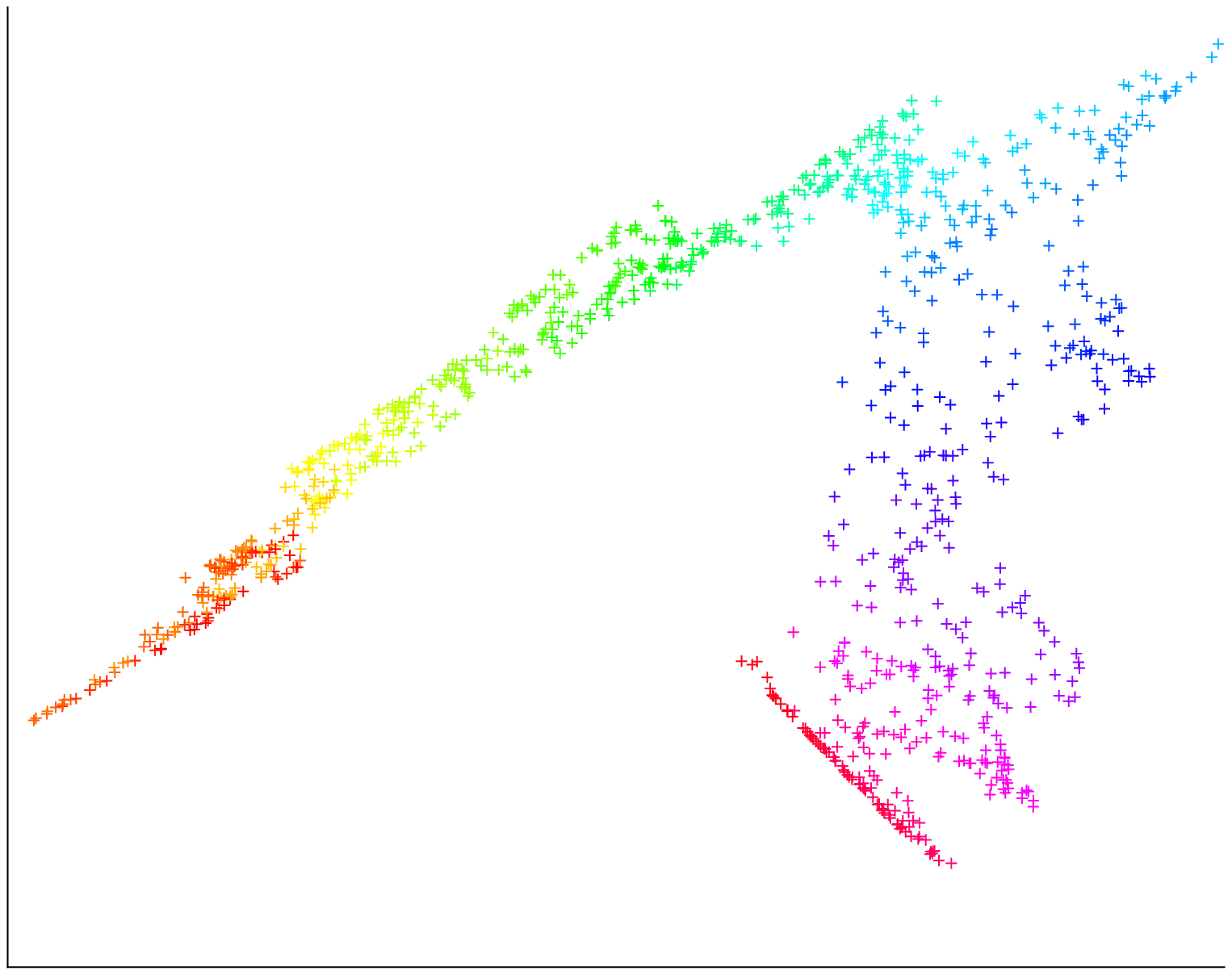}
		\centerline{(d)}
	\end{minipage}
	\vspace{1mm}
	\begin{minipage}[b]{.24\textwidth}
		\centering
		\includegraphics[width=4.8cm]{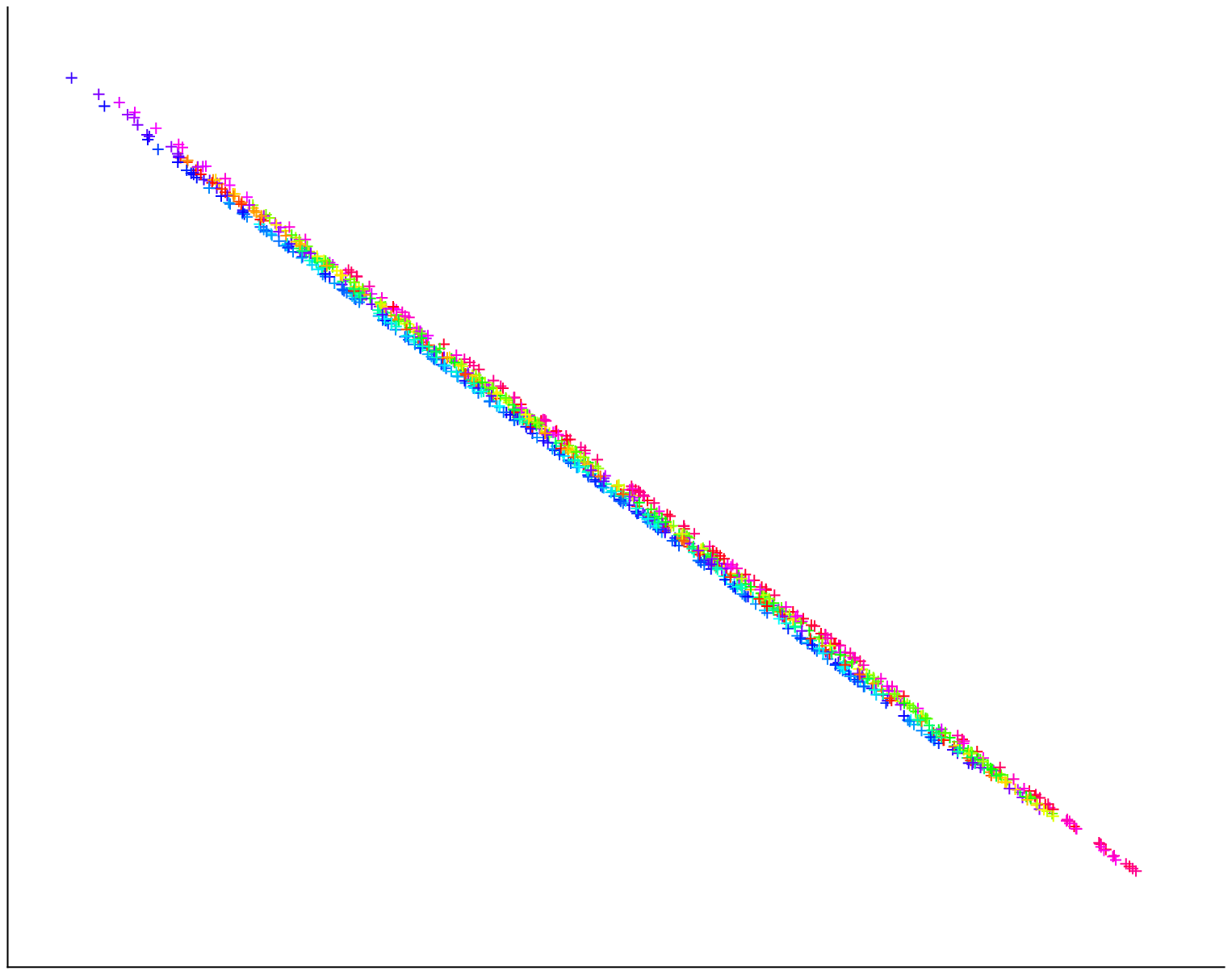}
		\centerline{(e)}
	\end{minipage}
	\vspace{1mm}
	\begin{minipage}[b]{.24\textwidth}
		\centering
		\includegraphics[width=4.8cm]{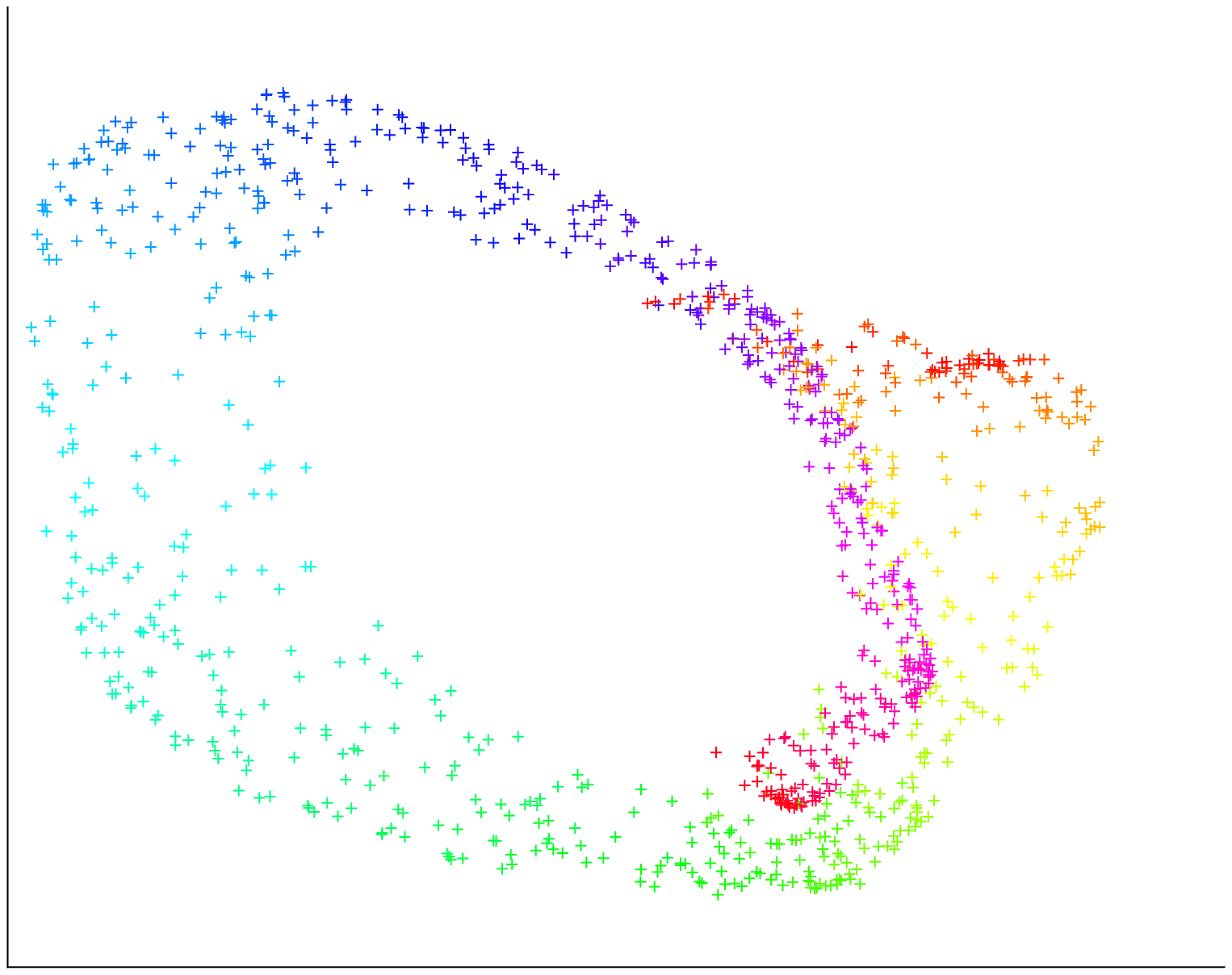}
		\centerline{(f)}
	\end{minipage}
	\vspace{1mm}
	\caption{{Embedding results of two manifolds: linear hyperplane and trefoil (a) visualization of $\{\bby_i\}_{i=1}^N$; and $\{\bbpsi_i\}_{i=1}^N$ obtained from (b)  PCA; (c) LLE with $K=20$; (d) LNEG with $K=20$; (e) LLE with $K=40$; (f) LNEG with $K=40$.} }
	\label{fig:swissroll}
\end{figure}

\subsection{Local Nonlinear Embedding}
Algorithm~\ref{algo:lneg} is tested using $\bbK_y$ as in \eqref{eq:Kx} for the locally nonlinear embedding (LNE) without and with graph regularization (the latter abbreviated as LNEG), both also compared with LLE and PCA. For all experiments, the graph $\mathcal{G}$ is constructed with adjacency matrix $\bbA$ with $(i,j)$th entry $a_{ij} = \bm{y}_i^\top\bm{y}_j/\|\bm{y}_i\|\|\bm{y}_j\|$. Two types of tests are carried out in order to: a) evaluate embedding performance for a single manifold; and b) assess how informative the low-dimensional embeddings are for distinguishing different manifolds.

\begin{figure*}[t]
	\begin{minipage}[b]{.24\textwidth}
		\centering
		\includegraphics[width=4.8cm]{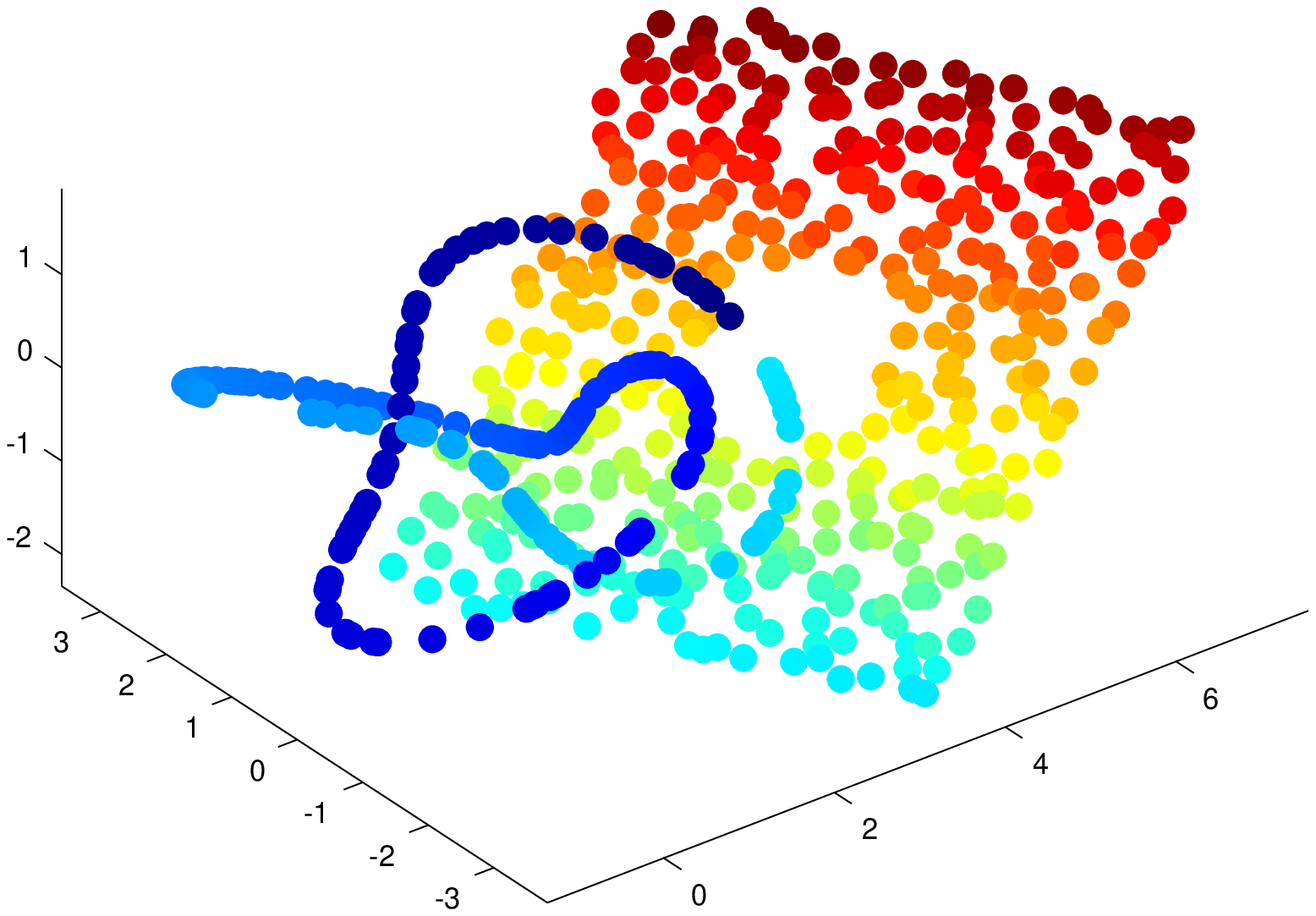}
		\centerline{(a)}
	\end{minipage}
	\begin{minipage}[b]{.24\textwidth}
		\centering
		\includegraphics[width=4.8cm]{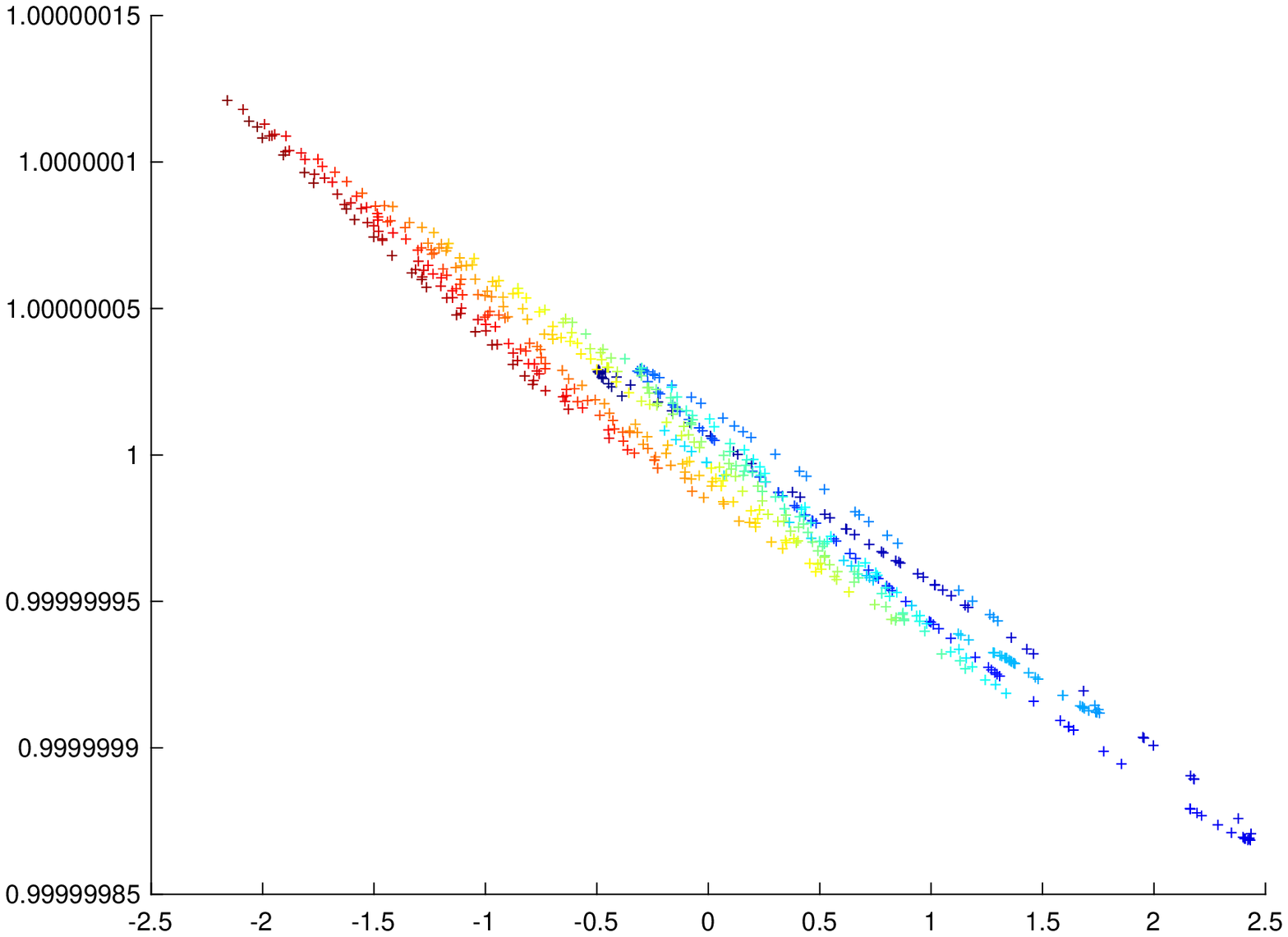}
		\centerline{(b)}
	\end{minipage}
	\begin{minipage}[b]{.24\textwidth}
		\centering
		\includegraphics[width=4.8cm]{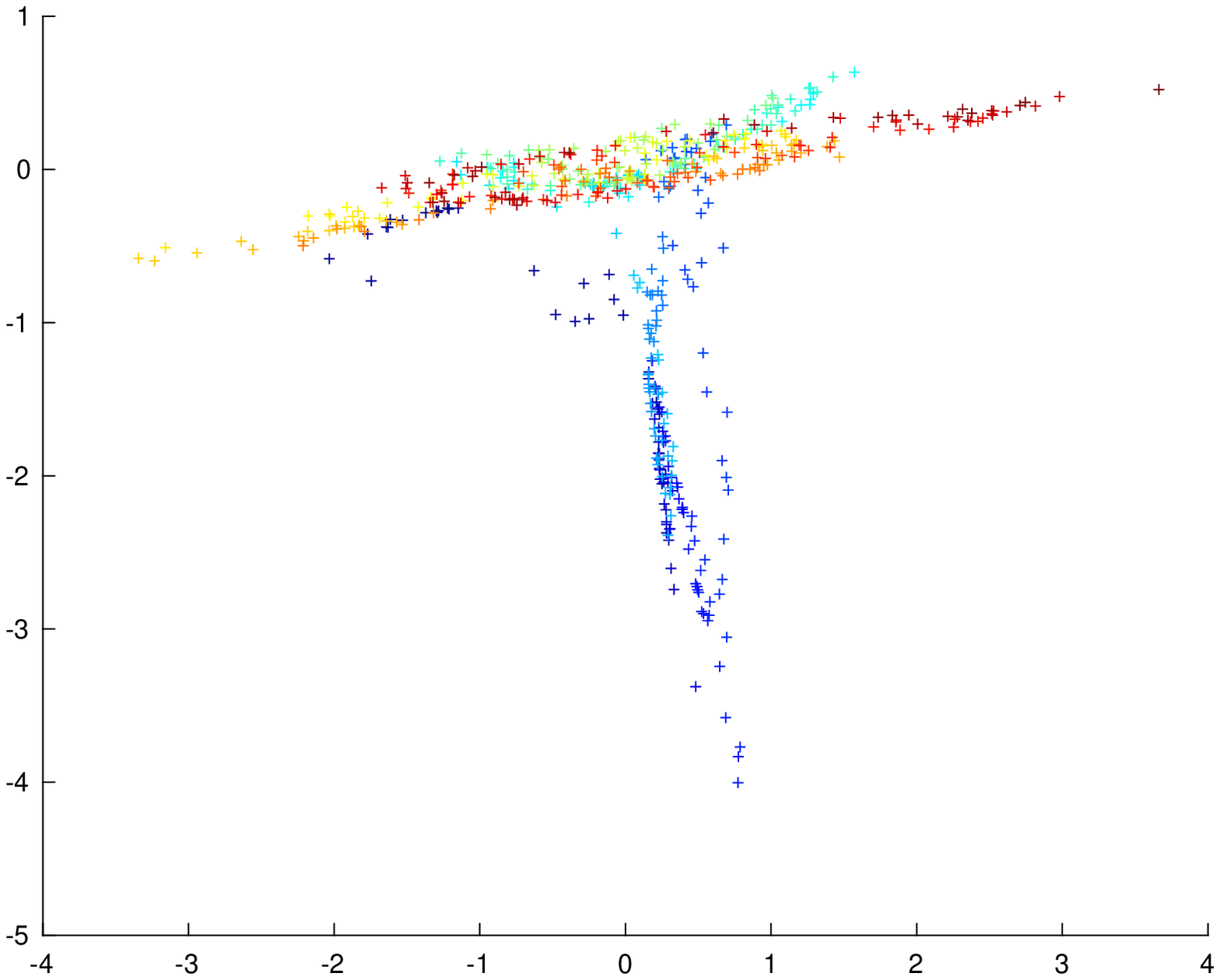}
		\centerline{(c)}
	\end{minipage}
	\begin{minipage}[b]{.24\textwidth}
		\centering
		\includegraphics[width=4.8cm]{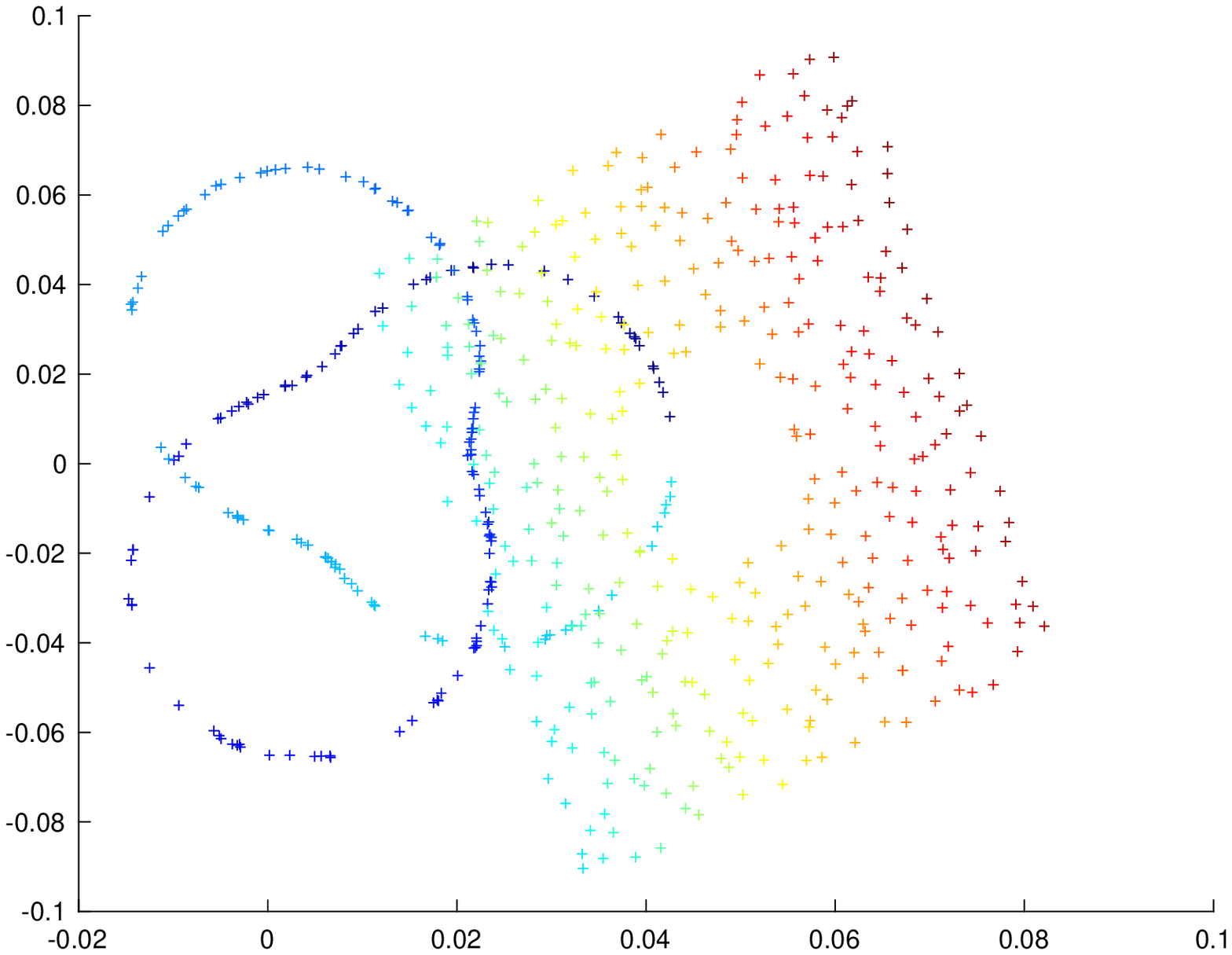}
		\centerline{(d)}
	\end{minipage}
	\vspace{2mm}
	\caption{{Embedding results of two manifolds: a linear hyperplane with hole and a trefoil (a) visualization of two manifolds; and $\{\bbpsi_i\}_{i=1}^N$ obtained from (b) LLE with $K=40$; and, (c) LNEG with $K=40$ and $P=2$; (d) PCA.} }
	\vspace{3mm}
	\label{fig:plane-trefoil}
\end{figure*}
\begin{figure*}[t]
	\begin{minipage}[b]{.24\textwidth}
		\centering
		\includegraphics[width=4.8cm]{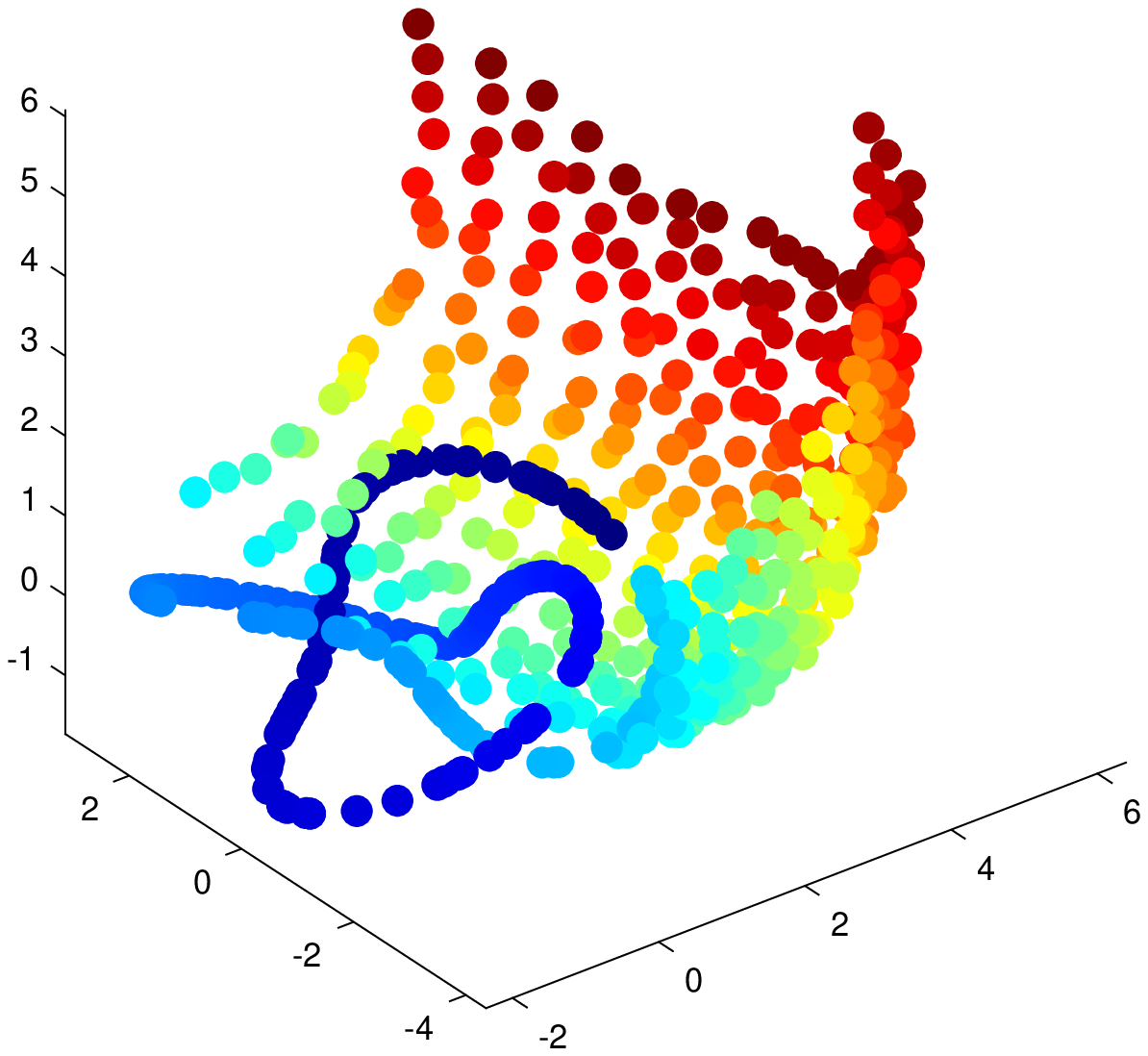}
		\centerline{(a)}
	\end{minipage}
	\begin{minipage}[b]{.24\textwidth}
		\centering
		\includegraphics[width=4.8cm]{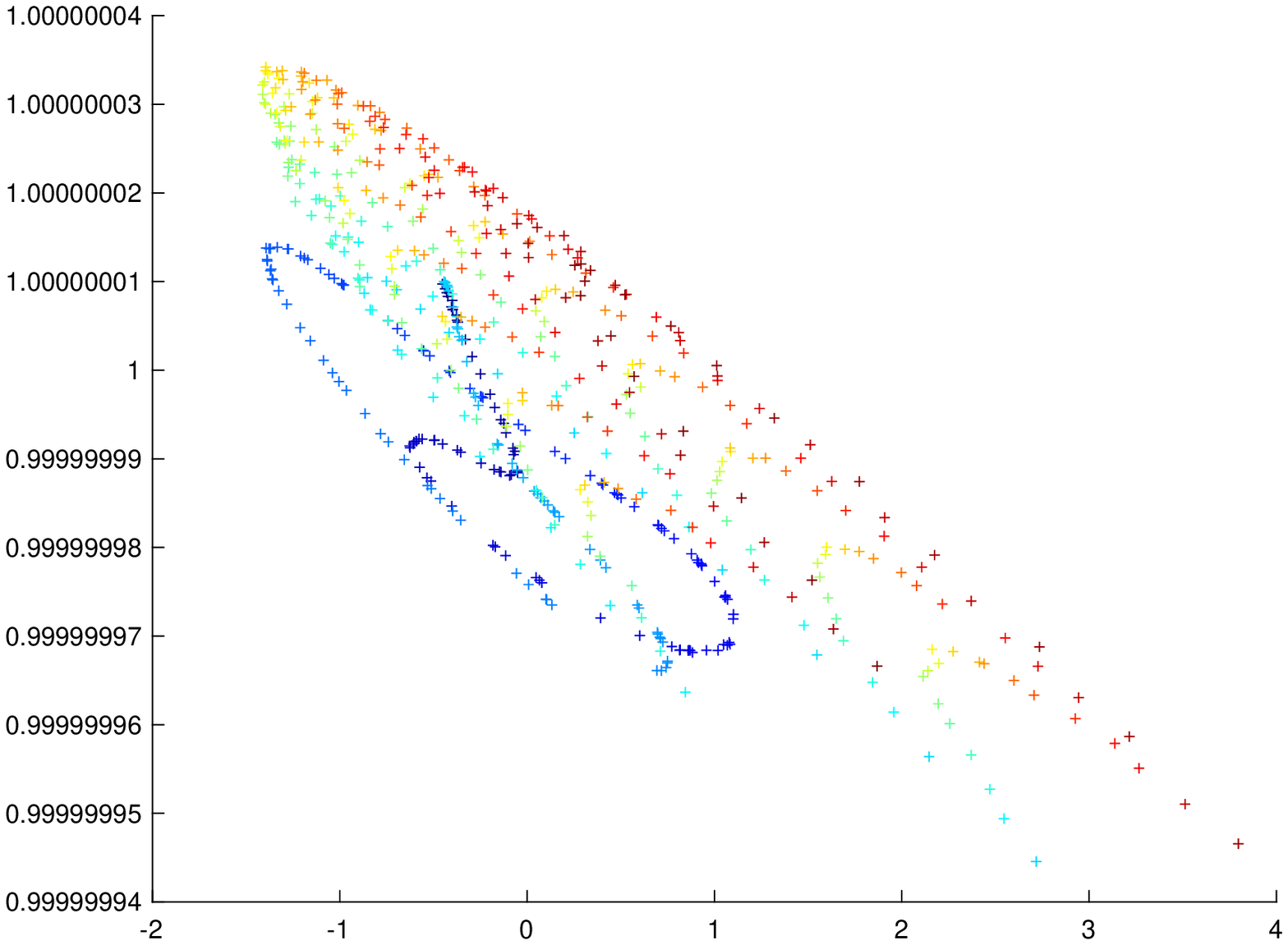}
		\centerline{(b)}
	\end{minipage}
	\begin{minipage}[b]{.24\textwidth}
		\centering
		\includegraphics[width=4.8cm]{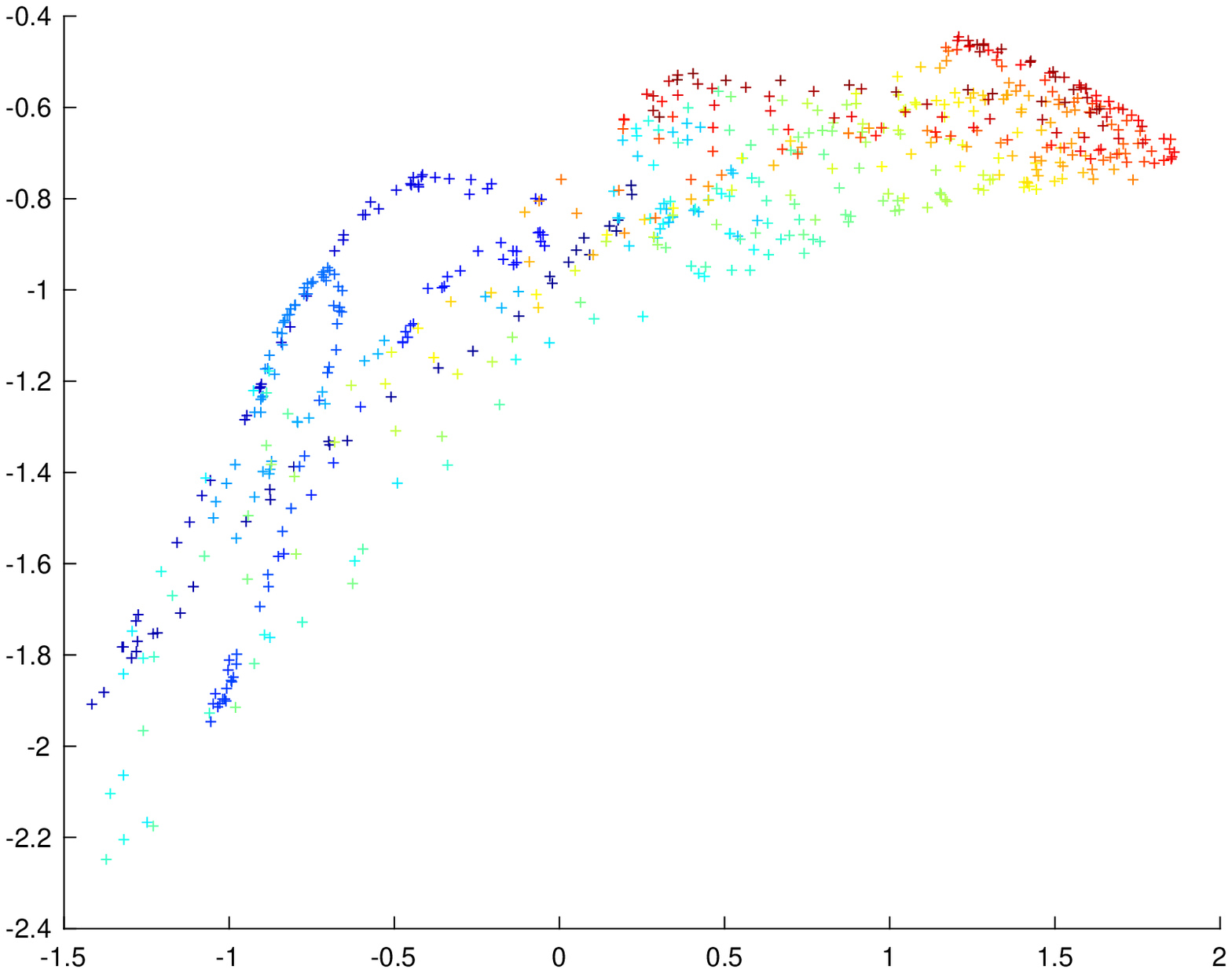}
		\centerline{(c)}
	\end{minipage}
	\begin{minipage}[b]{.24\textwidth}
		\centering
		\includegraphics[width=4.8cm]{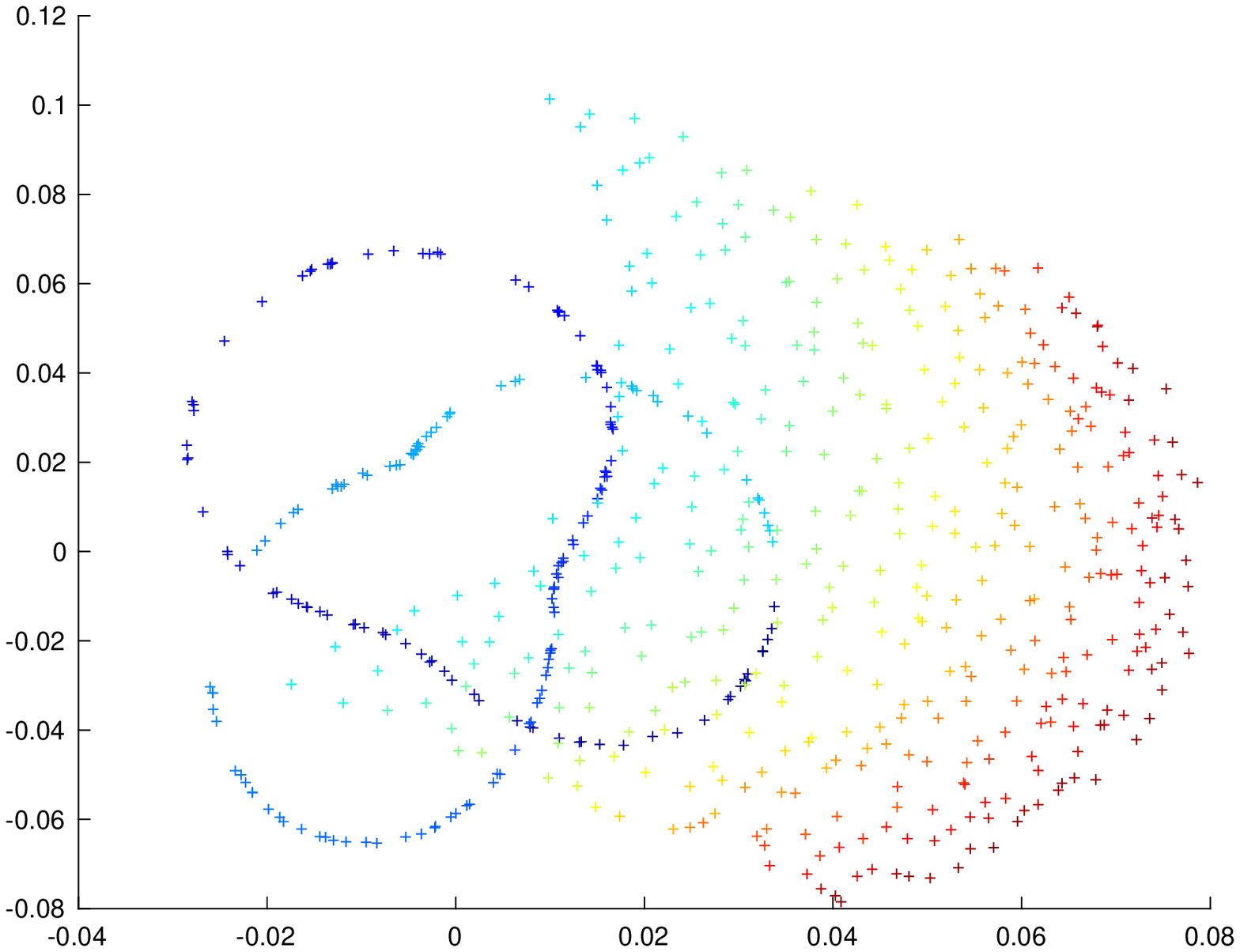}
		\centerline{(d)}
	\end{minipage}
	\vspace{3mm}
	\caption{{Embedding results of two manifolds: a nonlinear sphere and a trefoil (a) visualization of two manifolds; and $\{\bby_i\}_{i=1}^N$ obtained from (b) LLE with $K=40$; (c) LNEG with $K=40$ and $P=3$; and (d) PCA. } }
	\label{fig:sphere-trefoil}
\end{figure*}

\noindent\textbf{Embedding experiment.} In this experiment, we test the embedding performance of the proposed method. A 3-dimensional Swiss roll manifold is generated, and $1,000$ data are randomly sampled from the manifold as shown in Figure~\ref{fig:swissroll} (a). Figure \ref{fig:swissroll} (b) illustrates the $2$-dimensional embeddings obtained from PCA, while Figs. \ref{fig:swissroll}~(c) and (d) illustrate the resulting embeddings from LLE and LNEG respectively, where neighborhoods of $k=20$ data are considered. Figs. \ref{fig:swissroll}~(e) and (f) depict embeddings obtained by LLE and LNEG with $k=40$. The regularization parameter of LNEG is set to $\gamma=0.1$, and the polynomial order is set to $P=2$. Clearly, by exploiting the nonlinear relationships between data, the resulting low-dimensional representations are capable of better preserving the structure of the manifold, thus allowing for more accurate visualization.


\begin{table*}
	\begin{center}
		\vspace{0.3cm}
		\begin{tabular}[t]{ c || c|c  |c||c|c|c }
			\hline
			& \multicolumn{3}{c||}{Plane-hole-trefoil} &\multicolumn{3}{c}{Sphere-trefoil}\\ \hline 
			$K$ neighbours & \textbf{LLE} &\textbf{LNE}& {\bf LNEG}   & {\bf LLE} &{\bf LNE}& {\bf LNEG}  \\ \hline 
			5 & 0.25 &0.18 & 0.18  &0.29&0.21&0.17\\ \hline
			10&  0.44 & 0.21& 0.18 &0.39&0.27&0.16\\ \hline
			20&  0.15&0.13 & 0.17&0.48&0.26&0.14\\ 
			\hline
			30 & 0.21 & 0.20 & 0.17&0.46&0.28&0.19\\ \hline
			40 & 0.36 & 0.20 & 0.17&0.39&0.21&0.20\\ \hline
			\textbf{PCA} & \multicolumn{3}{c||}{0.49} &\multicolumn{3}{c}{0.43}\\ \hline
		\end{tabular}
	\end{center}
	\caption{Clustering error rate on low-dimensional representations obtained from: LLE, LNE, LNEG and PCA.}\label{tab:1}
\end{table*}

\noindent\textbf{Clustering experiment.} In this experiment, the ability of  Algorithm~\ref{algo:lneg} to provide meaningful embeddings for clustering of different manifolds is assessed. Two 3-dimensional manifolds, a linear hyperplane with a hole around the origin and a trefoil are generated on the same ambient space as per~\cite{smce}, and $200$ and $400$ data are sampled from them. Here each manifold corresponds to a different cluster. Figure~\ref{fig:plane-trefoil}(a) illustrates the sampled points from the generated manifolds. Matrices $\bbZ_1\in\mathbb{R}^{3\times 200}$ and $\bbZ_2\in\mathbb{R}^{3\times 400}$ contain the data generated from the linear hyperplane and the trefoil. Both manifolds are then linearly embedded in $\mathbb{R}^{100}$, that is $\bbY_i=\bbP\bbZ_i+\bbE_i$, where $\bbP\in\mathbb{R}^{100\times 3}$ is an orthonormal matrix, and $\bbE$ is a noise matrix with entries sampled from a zero mean Gaussian distribution with variance $0.01$. Afterwards, the $100$-dimensional data in $\bbY:=[\bbY_1~ \bbY_2]$ are embedded into $2$-dimensional representations $\bbPsi\in\mathbb{R}^{2\times 600}$ using LLE, LNEG and PCA. Figures. \ref{fig:plane-trefoil}(b), (c), and (d) depict the $2$-dimensional embeddings $\bbPsi$ provided by LLE, LNEG, and PCA, respectively. Similarly, Figure~\ref{fig:sphere-trefoil} illustrates the resulting embeddings when $\bbZ_2$ is sampled from a nonlinear sphere. In both cases, the nonlinear methods result in embeddings that separate the two manifolds. To further assess the performance, $K$-means is carried out on the resulting $\bbPsi$~\cite{lloydkmeans}. 
Table~\ref{tab:1} shows the clustering error when running $K$-means on the low-dimensional embeddings given by PCA, LLE, LNE and LNEG, across different values of $k$. 
The proposed approaches provide embeddings that enhance separability of the two manifolds, resulting in lower clustering error compared to LLE and PCA. In addition, greater performance gain is observed when both manifolds are nonlinear, as in the case of Figure~\ref{fig:sphere-trefoil}. The graph regularized method performs slightly better than that without regularization.

\section{Conclusions}
\label{sec:conclusion}
This paper introduced a general framework for nonlinear dimensionality reduction over graphs. By leveraging nonlinear relationships between data, low-dimensional representations were obtained to preserve these nonlinear correlations. Graph regularization was employed to account for additional prior knowledge when seeking the low-dimensional representations. An efficient algorithm that admits closed-form solution was developed along with a multi-kernel based algorithm that can handle settings where the nonlinear relationship between data is unknown. Furthermore, pertinent generalizations of the proposed schemes were provided. Several tests were conducted on simulated and real data to demonstrate the effectiveness of the proposed approaches. To broaden the scope of this study, several intriguing directions open up:  a) online implementations that can handle streaming data; and b) generalizations to cope with large-scale graphs and high-dimensional datasets.
\appendices

\section{Proof of \eqref{eq:dPCA3}}
\label{app:kernel}
Consider the objective function of \eqref{eq:dPCA3}, and define $\bbB:=\bbPsi^\top \bbPsi$. Then \eqref{eq:dPCA3} can be rewritten as
%
\begin{align}
\label{eq:app1-2}
&\min_{\bbB:{\rm rank}(\bbB)=d}\|\bbY^\top\bbY -\bbB\|_F^2
\end{align}
where the $\text{rank}(\bbB) = d$ constaint comes from the fact that $\bbB=\bbPsi^\top \bbPsi$ and $\bbPsi$ is a $d\times N$ matrix with $d\leq N$. The optimal solution $\bbB^{*}$ of \eqref{eq:app1-2} is given by the $d$ leading singular values and corresponding singular vectors of $\bbY^{\top} \bbY$ \cite{saad1992numerical}. Since $\bbY = \bbU\bbSigma\bbV^{\top}$ we have $\bbY^{\top}\bbY = \bbV\bbSigma^2\bbV^{\top}$, and consequently

\begin{align}
\label{eq:app1-3}
{\bbB}^{*}={\bbV}_d\bbSigma_d^{2}\bbV_d^\top
\end{align}
where $\bbV_d$ is a sub-matrix of $\bbV$ containing the $d$ singular vectors corresponding to the leading $d$ eigenvalues. It follows from \eqref{eq:app1-3} and $\bbB = \bbPsi^{\top}\bbPsi$ that
\begin{equation}
\bbPsi = \bbSigma_d\bbV_d^{\top}
\end{equation} 
which is the low-dimensional representation matrix provided by dual PCA [cf.~\eqref{eq:dPCA2}].
To complete the proof, just recall that
\begin{align}
\bbPsi\bbPsi^\top=\bbSigma_d^2=\bbLambda_d
\end{align}
where $\bbLambda_d$ contains the leading $d$ eigenvalues of $\bbB^{*}$.

\section{Proof of \eqref{eq:theta}}
\label{app:theta}
Here, we will show that having found $\mathbf \Psi$, the coefficients $\{\theta_q\}$ in \eqref{eq:mkl} can be obtained as in \eqref{eq:theta}.  
 Specifically, when the $\bbPsi$ is available, $\bbtheta$ can be obtained by
\begin{align}
\label{eq:app1}
\min_{\bm{\theta}}~~ &-\text{tr}(\bbPsi(\sum_{q=1}^Q\theta_q\bbK_y^{(q)})\bbPsi^\top)\nonumber\\
&\text{s.t.} ~~\|\bbtheta\|_2^2\leq 1,~~~\bbtheta\geq\mathbf{0}. 
\end{align}
The Lagrangian of \eqref{eq:app1} is 
\begin{align}
\label{eq:app2}
{\cal L}(\bbtheta, \lambda)=\text{tr}(\bbPsi(\sum_{q=1}^Q\theta_q\bbK_y^{(q)})\bbPsi^\top)+\lambda(\bbtheta^\top\bbtheta-1).
\end{align}
where $\lambda>0$ is the Lagrange multiplier. Taking the gradient of ${\cal L}(\bbtheta, \lambda)$ with respect to $\theta_q$ and equating  it to zero we have
\begin{align}
\label{eq:app3}
-\text{tr}(\bbPsi \bbK_y^{(q)}\bbPsi^\top) +\lambda \theta_q=0, ~~\forall q=1,\dots, Q
\end{align}
which yields
\begin{align}
\label{eq:app4}
\theta_q=\frac{1}{\lambda} \text{tr}(\bbPsi \bbK_y^{(q)}\bbPsi^\top).
\end{align}
Taking the gradient of ${\cal L}(\bbtheta, \lambda)$ with respect to $\lambda$ and setting it to $0$ we obtain
\begin{align}
\label{eq:app5}
\sum_{q=1}^Q \theta_q^2=1.
\end{align}
Substituting \eqref{eq:app4} into \eqref{eq:app5}, we arrive at
\begin{align}
\label{eq:app6}
\lambda=\sqrt{\sum_{q=1}^Q (\text{tr}(\bbPsi \bbK_y^{(q)}\bbPsi^\top))^2 }. 
\end{align}
Combining \eqref{eq:app4} with \eqref{eq:app6} leads to \eqref{eq:theta}.

\newpage
\balance
\bibliography{net,myabrv}
\bibliographystyle{IEEEtranS}
\end{document}